\documentclass{article} %
\usepackage{iclr2025_conference,times}

\usepackage{amsmath,amsfonts,bm}

\def\eqref#1{equation~\ref{#1}}

\def\1{\bm{1}}

\DeclareMathAlphabet{\mathsfit}{\encodingdefault}{\sfdefault}{m}{sl}
\SetMathAlphabet{\mathsfit}{bold}{\encodingdefault}{\sfdefault}{bx}{n}

\usepackage{arydshln}

\usepackage{times}
\usepackage{latexsym}
\usepackage{enumitem}

\usepackage[T1]{fontenc}

\usepackage[utf8]{inputenc}

\usepackage{microtype}

\usepackage{inconsolata}

\usepackage{multirow}
\usepackage{multicol}
\usepackage{amsmath}
\usepackage{amssymb}
\usepackage{breqn}
\usepackage{amsfonts}
\usepackage{algorithm}
\usepackage{algpseudocode}
\usepackage{wrapfig}

\usepackage{booktabs} %
\usepackage{xspace}
\definecolor{azure}{rgb}{0.0, 0.5, 1.0}
\PassOptionsToPackage{hyphens}{url}
\usepackage[backref=page]{hyperref}
\hypersetup{
     colorlinks   = true,
     citecolor    = cyan,
     linkcolor    = azure,
     urlcolor     = cyan,
}
\usepackage[capitalise,noabbrev]{cleveref}
\usepackage{subcaption}
\usepackage{multirow}
\usepackage{multicol}
\usepackage{color,xcolor}
\usepackage{booktabs}
\usepackage{adjustbox}
\usepackage{supertabular}
\usepackage{stmaryrd}
\usepackage{makecell}
\usepackage{marvosym}
\usepackage{colortbl}
\usepackage[framemethod=TikZ]{mdframed}
\usepackage[most]{tcolorbox}

\usepackage{etoolbox}
\makeatletter
\patchcmd{\BR@backref}{\newblock}{\newblock(page~}{}{}
\patchcmd{\BR@backref}{\par}{)\par}{}{}
\makeatother

\newcolumntype{g}{>{\columncolor{tbgray}}c}

\definecolor{red}{rgb}{0.8, 0.0, 0.0}
\definecolor{green}{rgb}{0.0, 0.5, 0.0}
\definecolor{tbgray}{gray}{.92}
\definecolor{pastelLavender}{rgb}{0.5, 0.4, 0.6} %
\definecolor{cgray}{gray}{.8}

\usepackage{listings}
\usepackage{tikz}
\newcommand{\pgftextcircled}[1]{
    \setbox0=\hbox{#1}%
    \dimen0\wd0%
    \divide\dimen0 by 2%
    \hspace{-0.2em}\begin{tikzpicture}[baseline=(a.base)]%
        \useasboundingbox (-\the\dimen0,0pt) rectangle (\the\dimen0,0pt);
        \node[circle,draw=cgray,outer sep=0pt,inner sep=0.05ex](a){\scriptsize#1};
    \end{tikzpicture}\hspace{-0.2em}
}

\newcommand{\sys}{{CtrlSynth}\xspace} %

\renewcommand{\cite}[1]{\citep{#1}}

\iclrfinalcopy

\title{CtrlSynth: Controllable Image Text Synthesis for Data-Efficient Multimodal Learning}

\author{Qingqing Cao \& Mahyar Najibi \\ Apple \And Sachin Mehta \\ Meta \thanks{Work done while at Apple.}
}

\definecolor{codegreen}{rgb}{0,0.6,0}
\definecolor{codegray}{rgb}{0.5,0.5,0.5}
\definecolor{codepurple}{rgb}{0.58,0,0.82}
\definecolor{backcolour}{rgb}{0.95,0.95,0.92}

\definecolor{tablegreen}{rgb}{0.9,0.99,0.9}
\definecolor{tablered}{rgb}{0.99,0.9,0.9}

\begin{document}

\maketitle

\begin{abstract}
    Pretraining robust vision or multimodal foundation models (\eg, CLIP) relies on large-scale datasets that may be noisy, potentially misaligned, and have long-tail distributions. Previous works have shown promising results in augmenting datasets by generating synthetic samples. However, they only support domain-specific ad hoc use cases (\eg, either image or text only, but not both), and are limited in data diversity due to a lack of fine-grained control over the synthesis process. 
    In this paper, we design a \emph{controllable} image-text synthesis pipeline, \sys, for data-efficient and robust multimodal learning. The key idea is to decompose the visual semantics of an image into basic elements, apply user-specified control policies (\eg, remove, add, or replace operations), and recompose them to synthesize images or texts. The decompose and recompose feature in \sys allows users to control data synthesis in a fine-grained manner by defining customized control policies to manipulate the basic elements. \sys leverages the capabilities of pretrained foundation models such as large language models or diffusion models to reason and recompose basic elements such that synthetic samples are natural and composed in diverse ways. \sys is a closed-loop, training-free, and modular framework, making it easy to support different pretrained models. With extensive experiments on 31 datasets spanning different vision and vision-language tasks, we show that \sys substantially improves zero-shot classification, image-text retrieval, and compositional reasoning performance of CLIP models.
\end{abstract}

\section{Introduction}

High-quality large-scale datasets have driven the success of large foundational AI models~\citep{radfordLearningTransferableVisual2021,rombachHighResolutionImageSynthesis2022,touvronLlamaOpenFoundation2023}. Collecting and annotating datasets at large-scale is challenging and costly. One solution is to crawl data from the web; however, web data is noisy~\cite{laiVeCLIPImprovingCLIP2024,Kang_2023_ICCV}, has long-tail distributions~\citep{udandaraoNoZeroShotExponential2024}, and often causes privacy or copyright issues~\citep{schuhmann2022laionb}. Synthetic data presents a viable and complementary alternative to overcome these challenges, as it allows for precise control over data generation and customization to meet specific requirements. A large body of work has focused on improving the quality of synthetic data for image and text data, from the generation of high-quality images~\cite{dunlapDiversifyYourVision2023,islamDiffuseMixLabelPreservingData2024} to the improvement of synthetic captions~\cite{laiVeCLIPImprovingCLIP2024,fanImprovingCLIPTraining2023}. While these works have shown that synthetic data successfully improves model performance for various vision or vision-language tasks, their synthetic pipeline is often ad hoc and tailored to specific purposes such as training better CLIP models or improving domain-specific vision models~\citep[\eg, DiffuseMix uses diffusion models to augment images and improves accuracy on image classification tasks][]{islamDiffuseMixLabelPreservingData2024}. These data synthesis works also lack explicit fine-grained control over the generated texts or images, which are important for tasks with long-tail distribution (\eg, augmenting tail class samples) or enforcing safety requirements~\citep[\eg, mitigating biased or sensitive content generation][]{schramowskiSafeLatentDiffusion2023}. 

\begin{figure*}[t!]
	\begin{center}\vspace{-.5em}
		\includegraphics[width=\linewidth]{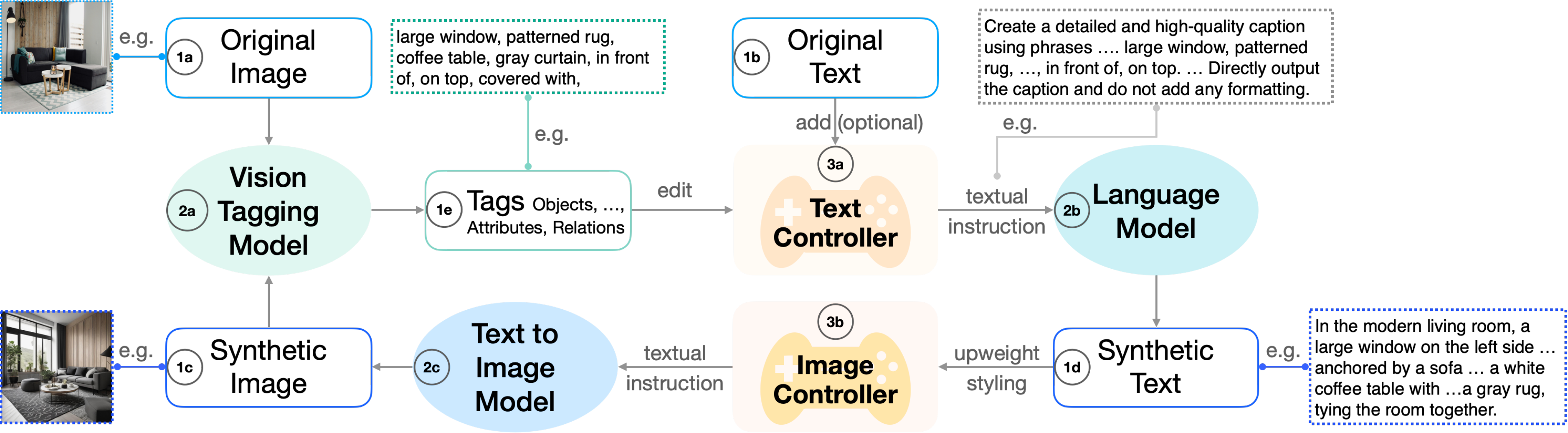}
  \vspace{-.3em}
		\caption{\sys: A modular, closed-loop, controllable data synthesis system. The \textit{oval nodes} indicate that the pretrained models and \textit{rounded boxes} represent text or image data. The text and image controllers are used to guide the data synthesis.}
		\label{fig:arch}
  \vspace{-1.5em}
	\end{center}
\end{figure*}

In this work, we aim to systematically control the synthetic pipeline for generating image-text data while accommodating different use cases (\eg, improving long-tail task performance, enhancing compositional reasoning of CLIP models, etc.). Our intuition is that large foundation models are already pretrained on a wide range of data and contain general knowledge about concepts, objects, and their relationships. For example, text-to-image models~\citep[\eg,][]{rombachHighResolutionImageSynthesis2022,podell2024sdxl} can generate detailed high-quality images based on text instructions. Similarly, large language models (LLMs) ~\citep[\eg,][]{ChatGPT,touvronLlamaOpenFoundation2023} have strong instruction-following capabilities, which can be used to control the text data generation. \sys leverages these large pretrained models to build a modular and controllable synthetic data generation pipeline. \sys allows users to apply explicit control instructions to guide data generation for images and texts. Unlike previous data synthesis works that use image-captioning models to directly generate captions given an image~\citep[\eg,][]{liWhatIfWe2024,laiVeCLIPImprovingCLIP2024}, \sys decomposes image-to-text generation process into two separate steps, providing more fine-grained control to users for synthesizing data. \cref{fig:arch} shows an overall architecture of the \sys pipeline. 
For an input image, \sys first uses a pretrained vision model to extract key objects, attributes, and their relations as visual tags. It then uses a text controller to create text synthesis instructions and guide the LLM to use visual tags to generate high-quality text outputs. Similarly, we devise an image controller that steers how the text prompts (or caption) can be used to guide the diffusion model to generate a desired image. Users can also feed the generated synthetic images into the tagging model again, forming a closed-loop data pipeline. Then users can start with synthetic or original images and texts and further generate more image-text pairs. The text and image controllers are modular, allowing users to control any part of the text or image generation process.

Compared to previous works, \sys provides three main benefits: (1) \textbf{Controllable synthesis}: \sys allows users to define policies on the visual tags or texts; enabling granular control over text and image generation; (2) \textbf{Closed-loop system}: \sys requires no additional training and can synthesize text from images and vice-versa using existing pretrained models. This closed-loop design additionally provides automatic filtering and verification capabilities to discard undesirable synthetic samples without manual or heuristics-based rules. (3) \textbf{Flexible and scalable}: \sys is modular and allows users to change its components (\eg, pretrained models) easily. %
(4) \textbf{Extensive empirical results} We evaluate the effectiveness of \sys on different tasks (\eg, image classification, image-text retrieval, compositional reasoning, and long-tail tasks), covering \textbf{31 datasets} for vision and vision-language domains. We observe that \sys generated data improves the accuracy by (a) 23.4\% on retrieval tasks, (b) 5\% on the SugarCrepe compositional reasoning benchmark, and (c) 16\% $\sim$ 21\% for long-tail vision tasks.

\section{Related Work}

{\noindent \bf Data-Efficient Vision-Language Representation Learning.} Contrastive Language-Image Pretraining (CLIP)~\citep{radfordLearningTransferableVisual2021} has popularized visual representation learning from image-text pairs due to its strong zero-shot transfer capabilities. Many recent works have focused on improving the data efficiency of training CLIP models. SLIP~\citep{muSLIPSelfsupervisionMeets2022} brings self-supervised learning into a multitask learning framework to improve CLIP performance. FLIP~\citep{liScalingLanguageImagePreTraining2023} masks out image patches during CLIP training, improving training efficiency and zero-shot accuracy over baselines. CLIPA~\citep{liInverseScalingLaw2023,liCLIPAv2ScalingCLIP2023} further improves over FLIP ideas and reduces the number of image text tokens by block and syntax masking for CLIP training and it significantly reduces the training costs of CLIP models. LiT~\citep{zhaiLiTZeroShotTransfer2022} freezes the image encoder in CLIP models and achieves strong zero-shot transfer for CLIP models using much fewer data samples. All these techniques focus on improving the training methods for CLIP models to enable better vision-language representations. \sys improves data augmentation for CLIP training by synthesizing diverse image text samples. Our method is orthogonal and could potentially benefit from these methods. 

{\noindent \bf Image-text Data Augmentation.} Much recent work aims to improve the caption quality of image-text pairs. For example, VeCLIP~\citep{laiVeCLIPImprovingCLIP2024}, LaCLIP~\citep{fanImprovingCLIPTraining2023}, and ReCap~\citep{liWhatIfWe2024} leverage LLMs to synthesize new captions that are more informative and contain rich descriptions about the image. The key difference of \sys is that we provide more diverse and high-quality captions that outperform prior works (we will show in \cref{tab:zs-velip} and \cref{tab:zs-laclip}). This is because \sys breaks down the image semantics to allow more fine-grained control and recombination using LLM. 
Another line of work uses text-to-image models like diffusion models to generate synthetic images and augment downstream vision tasks. ALIA~\citep{dunlapDiversifyYourVision2023} uses language to guide the image editing process and provides domain-specific diversity to augment the image samples. DiffuseMix~\citep{islamDiffuseMixLabelPreservingData2024} augments image samples using diffusion models to blend original and generated images. EDA~\citep{trabuccoEffectiveDataAugmentation2023} generates variations of real images using diffusion models to maintain the semantics while augmenting image samples. These semantic image augmentation methods provide strong performance improvements on various vision datasets. Our \sys instead unifies the image and text synthesis via a closed-loop pipeline, it provides more flexibility and diverse synthetic samples while allowing more fine-grained control over the sample generation process.%

\section{\sys}
\label{sec:method}

\sys leverages semantic knowledge and reasoning skills of pretrained foundation models (\eg, large language and diffusion models) to generate diverse synthetic data samples in a controlled manner. Specifically, \sys consists of three foundation models: (1) a vision tagging model, (2) a large language model, and (3) a text-to-image model; plus the two text and image controllers. For a given real (\pgftextcircled{1a} in \cref{fig:arch}) or synthetic (\pgftextcircled{1c}) input image, a \emph{vision tagging model} (\pgftextcircled{2a}) extracts visual tags (\eg, objects, attributes, and their relationships) (\pgftextcircled{1e}). These tags describe the image's visual concepts and semantic contexts. The \emph{text controller} (\pgftextcircled{3a}) takes the image tags and user-defined control policies as inputs and generates instructions for synthesizing new text. An example control policy is to edit the tags or optionally add the text (\pgftextcircled{1b}) associated with the image. 
A \emph{large language model} (\pgftextcircled{2b}) then follows the instructions and generates the synthetic text (\pgftextcircled{1d}). The \emph{image controller} (\pgftextcircled{3b}) operates on the given input text 
and applies user-defined image control policies to output instructions for image synthesis. An example policy is to specify the style for generating artistic, cinematic, or realistic images. 
A \emph{text-to-image model} (\pgftextcircled{2c}) takes an image synthesis instruction provided by the image controller as an input and produces a synthetic image as an output (\pgftextcircled{1c}). 

\subsection{Key Components}
\label{sec:components}

\mdfdefinestyle{mdfexample1}{innertopmargin=0.2em,innerbottommargin=0.2em,%
innerleftmargin=0.2em,innerrightmargin=0.2em,%
roundcorner=2pt,outerlinewidth=0.1,linecolor=blue!50,hidealllines=true}

{\noindent \bf Vision Tagging Model.}
The goal of a vision tagging model (VTM) is to extract the basic visual elements (or tags) of an image, including all objects or entities, attributes (\eg, color, shape, and size), and visual relations (\eg, interaction between objects). 

\begin{wrapfigure}{r}{0.4\linewidth}
\vspace{-2.8em}
\resizebox{.98\linewidth}{!}{
\begin{minipage}[t]{\linewidth}
\begin{mdframed}[style=mdfexample1]
\centering
\includegraphics[width=\linewidth]{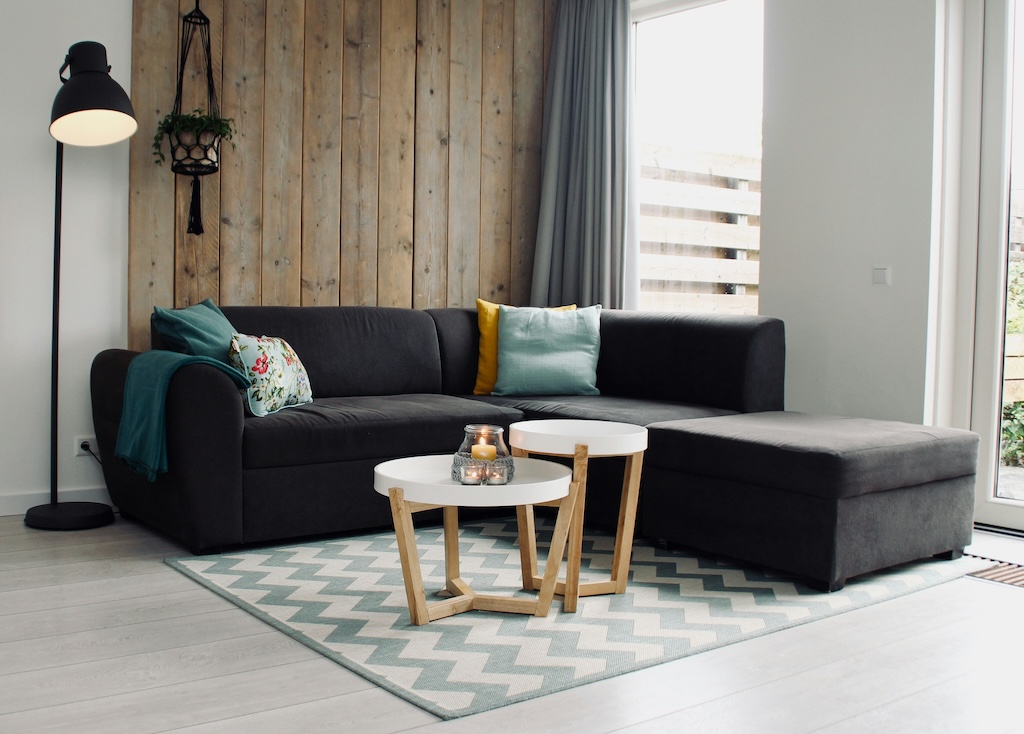}
\vspace{-2.em}
\caption*{\small
\textbf{Objects and attributes}: light candle, patterned rug, white coffee table, sectional sofa\\
\textbf{Relations}: in front of, on top, covered with \\
}
\vspace{-1.8em}
\caption{Visual tags of an example image\protect\footnotemark. Tags are non-exhaustive.}
\label{fig:vtm-ex}
\end{mdframed}
\end{minipage}
}
\vspace{-2.8em}
\end{wrapfigure}
\footnotetext{Image credit: \tiny \url{https://unsplash.com/photos/light-candle-on-round-white-coffee-table-and-sectional-sofa-GZ5cKOgeIB0}}

An example of extracting visual tags from VTM is shown in \cref{fig:vtm-ex}. The tagging model can be either a multi-label image classifier~\citep{mehtaCatLIPCLIPlevelVisual2024} that predicts diverse tags in the image, or a black box system (\eg an API service) that takes the input image and outputs the tags.

VTM, as a key component in \sys, can be a combination of an advanced captioning model~\citep{xiaoFlorence2AdvancingUnified2024} that generates comprehensive image descriptions and an LLM that extracts the visual tags from the captions to decompose the visual semantics of an image into a set of fine-grained visual concepts. \cref{sec:inference-details} includes more details about this hybrid VTM. %
These fine-grained visual concepts can be easily modified and recomposed to create new visual contexts. This decompose-recompose feature of vision tags provides a large control space for synthesizing diverse texts. 

Existing caption rewriting works (\eg,  VeCLIP~\cite{laiVeCLIPImprovingCLIP2024}) rely on a multimodal captioning model to generate captions that are short sentences containing visual concepts. Image captions can be very descriptive but often only cover the most salient object of the scene, they are coarse-grained in structure (whole sentence or paragraph), and are hard to modify. 
Our key distinction is that VTM produces a comprehensive list of metadata information that describes the visual concepts in an image as completely as possible.

{\noindent \bf Language Model.} Large language models (LLMs) have exhibited strong instruction-following capabilities. The goal of an LLM in \sys is to take an input textual instruction on how to generate a synthetic text that meets the requirements specified in the instruction. \sys employs the reasoning and composition capability of LLMs to recombine the visual image tags in the task instruction and compose new synthetic texts. The instruction for an LLM consists of three parts (\cref{fig:inst-ex}): \textit{(i) \textcolor{olive}{task template}} that specifies the details of the text synthesis task, \textit{(ii) \textcolor{teal}{task content}} that contains the actual visual tags (phrases) and an optional caption paired with the image, and \textit{(iii) \textcolor{brown}{task constraint}} that describes the style and formatting of the output text. Users can also apply custom policies over the instructions to guide the text synthesis process.   %

\begin{figure*}[t!]
    \centering
\vspace{-2.2em}
\begin{tcolorbox}[enhanced,opacityback=.4,
boxsep=1pt,left=5pt,right=5pt,top=6pt,bottom=6pt,colframe=white]
{\emph{\textcolor{olive}{Write a faithful caption by integrating the given phrases with the original sentence. Ensure any objects from the original caption are preserved while elaborating on the visual relationships and attributes provided in the phrases to create a more detailed depiction.} \textcolor{teal}{Given sentence: \{caption\}. Given phrases: \{phrases\}.} \textcolor{brown}{The caption should not contain any NSFW words. It should be grammatically correct. It should be concise, but not too short. Directly output the caption and do not add any formatting.}}}
\end{tcolorbox}
\vspace{-1.2em}
\caption{An example instruction for LLMs to synthesize texts.}
\label{fig:inst-ex}
\vspace{-1.8em}
\end{figure*}

{\noindent \bf Text-to-Image Model.} Text-to-image models generate novel and diverse image samples based on different input text prompts. \sys applies an image controller to account for the user-specified control policies and accordingly, updates the input text instructions from the previous step (i.e., language model). These updated instructions are then fed to text-to-image models for generating the image as an output. In our experiments, we use StableDiffusion models for text-to-image generation.

{\noindent \bf Text and Image Controllers.} The controller in \sys is a function that takes an input text and transforms it into a specific text instruction for the LLM or text-to-image model. 

The text controller accepts the visual tags of an image and a user-defined policy along with an optional original text as input and produces instructions to control the generation of synthetic text. %
In \sys, we study three predefined policies: (a) editing (remove, add, or replace) visual tags, (b) constraining the semantic meaning of a given sentence, and (c) styling the output text. Editing visual tags allows fine-grained control of synthetic visual content, for example, one can remove unwanted objects or attributes so they do not appear in the generated text. Constraining the meaning of synthetic text is useful in generating high-quality captions because many web-crawled captions are noisy. Enforcing the styling of output texts such as outputting into structured formats (\eg, JSON) makes the texts easier to use in downstream tasks. In our experiments, we use 10 example text control policies for synthesizing image captions (see \cref{sec:policy-details} for details).

The image controller is similar to the text controller in functionality. It mainly steers image generation via specific prompting. We study two simple control policies to show the controllability and utility of \sys. The first one involves weighting particular tags in the input prompt (lower or increase individual weights for a given tag) so that the output image has a different focus on the objects or attributes. The second policy applies different styles (\eg, cinematic, realistic, or art) to the output images for generating diverse content. %
Note that the control policies are flexible and can be easily modified for diverse use cases. For example, one can integrate more complex policies such as layout-guided~\citep{lianLLMgroundedDiffusionEnhancing2023} or planning-based~\citep{yangMasteringTexttoImageDiffusion2024} image generation.%

\subsection{Image Text Synthesis in \sys} 
\label{sec:synth}
\sys %
is a modular and closed-loop system by design and supports diverse image and text synthesis configurations. In this section, we first introduce different synthesis paths in \sys and then describe how the closed-loop feature allows \sys to filter out low-quality samples. 

\begin{figure*}[htb!]
    \centering
    \captionsetup[subfigure]{font=scriptsize}
  \subcaptionbox{Synthesis path $SP(1)$\label{fig:sp1}}%
  {\includegraphics[width=0.45\linewidth]{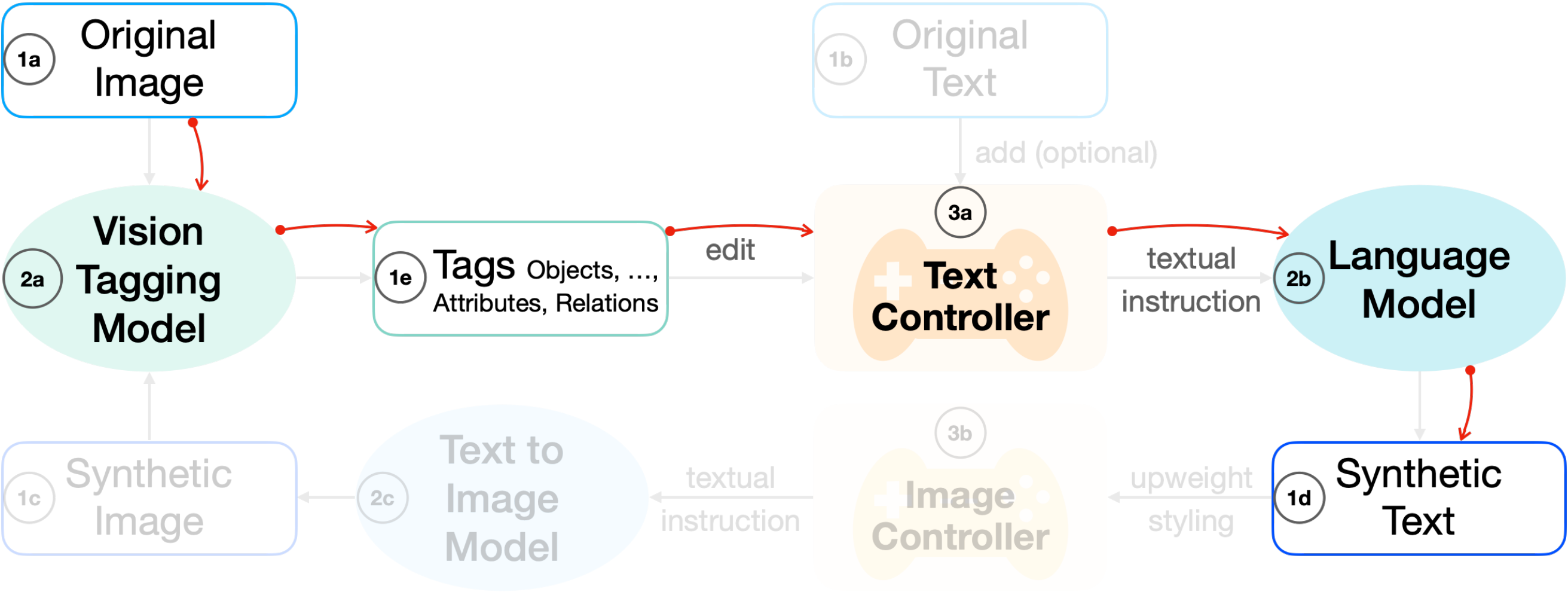}
  \vspace{-0.5em}
  }
  \hfill
  \subcaptionbox{Synthesis path $SP(2)$\label{fig:sp2}}%
  {\includegraphics[width=0.45\linewidth]{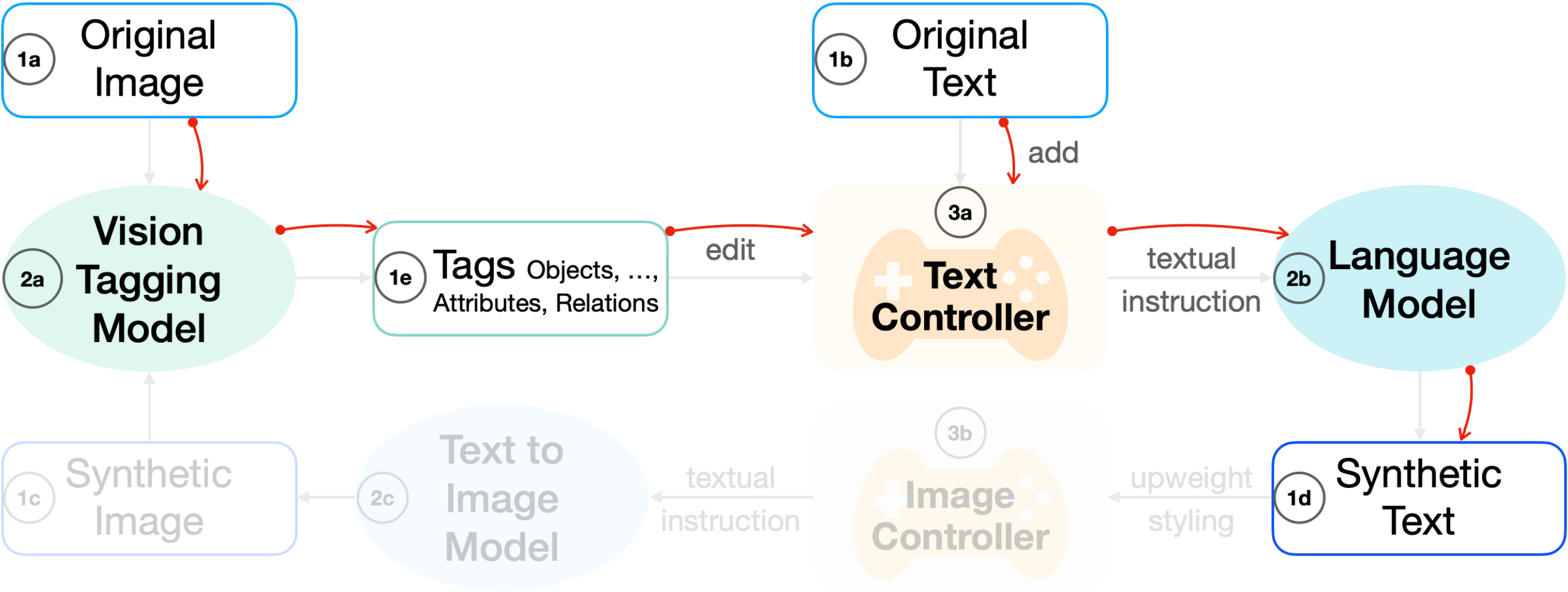}
  \vspace{-0.5em}
  }
   \hspace{\fill}\vspace{0.5em}
  \subcaptionbox{Synthesis path $SP(3)$\label{fig:sp3}}%
  {\includegraphics[width=0.45\linewidth]{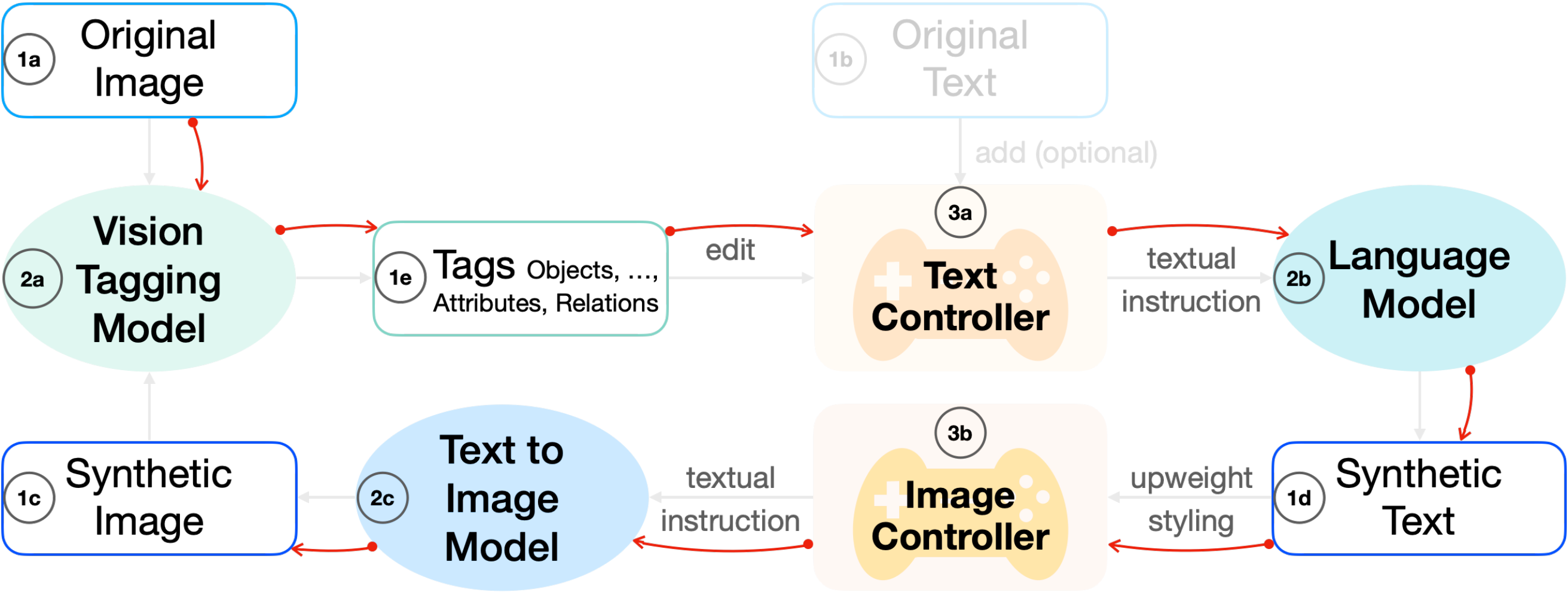}
  \vspace{-0.5em}
  }
  \hfill
  \subcaptionbox{Synthesis path $SP(4)$\label{fig:sp4}}%
  {\includegraphics[width=0.45\linewidth]{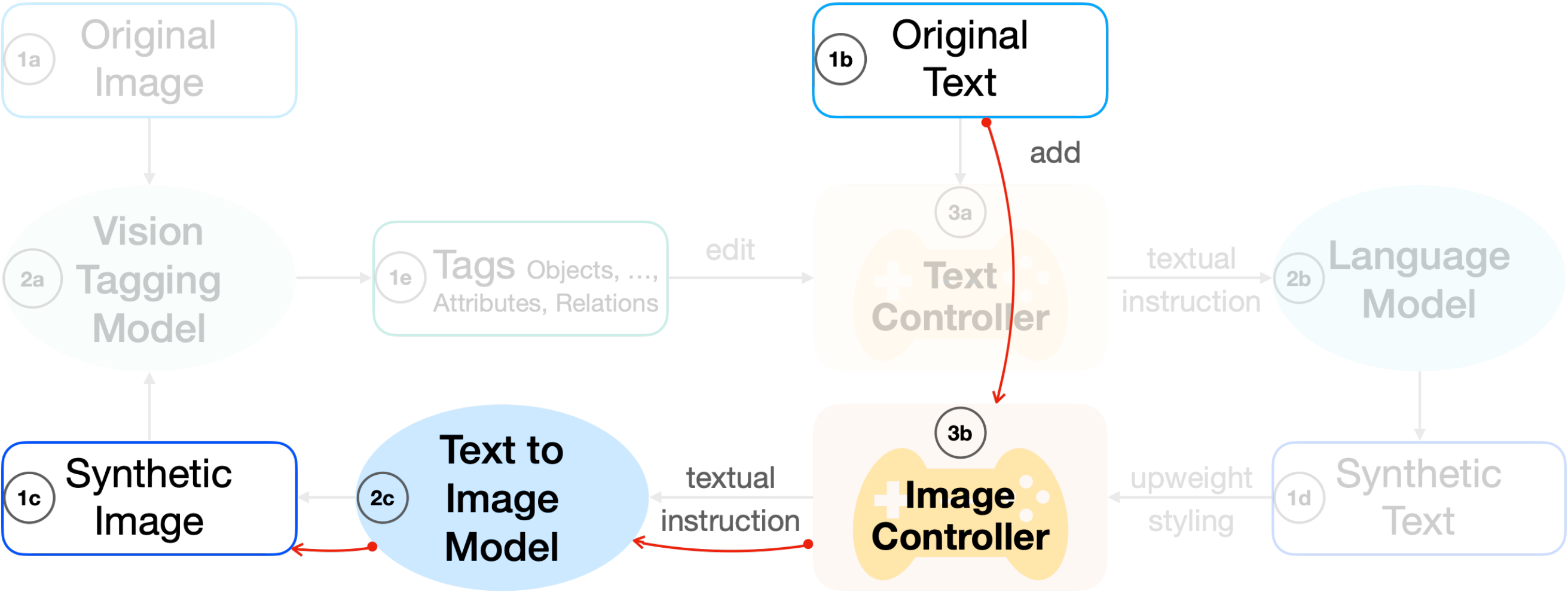}
  \vspace{-0.5em}
  }
 \hspace{\fill}
\vspace{-0.5em}
\caption{Different synthesis paths in \sys.}
\label{fig:sp}
\vspace{-1.em}
\end{figure*}

{\noindent \bf Flexible and diverse synthesis paths.} A data synthesis path ($SP$) starts and ends with a data node (rounded box in \cref{fig:arch}). We define the following synthesis paths: 

$SP(1)$: $1a \rightarrow 2a \rightarrow 1e \rightarrow 3a \rightarrow 2b \rightarrow 1d$. This path (\cref{fig:sp1}) means \sys generates a new text that describes the original image. The synthetic text $1d$ may not align with the semantics in the original image since the LLM can create new combinations of the visual tags and add information that does not exist in the image. Such new information provides useful semantic augmentation over the original image while containing similar visual concepts. %

$SP(2)$: $1a \rightarrow 2a \rightarrow 1e \xrightarrow{1b} 3a \rightarrow 2b \rightarrow 1d$. This path (\cref{fig:sp2}) is similar to the previous path but a key difference is that it constrains the synthetic text to be faithful\footnote{Or the opposite depending on the user-specified policy} to an original text. We can consider it as using the VTM and LLM to synthesize an improved text over the original one. We will show later in \cref{sec:ablation} that text samples generated from this path outperform previous works~\citep{laiVeCLIPImprovingCLIP2024,fanImprovingCLIPTraining2023} that rewrite noisy captions. We include the example prompts to reflect the control policies in \cref{sec:policy-details}. %

$SP(3)$: $1a \rightarrow 2a \rightarrow 1e \rightarrow 3a \rightarrow 2b \rightarrow 1d \rightarrow 3b\rightarrow 2c\rightarrow 1c$. This path (\cref{fig:sp3}) provides both synthetic text ($1d$) and image ($1c$) samples. $1c$ can be an effective image sample that augments the original image ($1a$) or can be paired with ($1d$) to augment the original image-text pair ($1a$ and $1b$).

$SP(4)$: $1b \rightarrow 3b \rightarrow 2c \rightarrow 1c$. This path (\cref{fig:sp4}) bypasses the language model and the original text is directly fed to the image controller and then generates a synthetic image ($1c$). The image sample could be a strong augmentation sample to the original image if the original text has a comprehensive and high-quality description.

Note that \sys supports more synthesis paths that are not listed above. For example, one can start with original text and use LLM to add creative elements and generate synthetic text and further use it to generate an image, i.e. $1b \rightarrow 3a \rightarrow 2b \rightarrow 1d \rightarrow 3b \rightarrow 2c \rightarrow 1c$. Another category of examples includes starting with synthetic texts or images and creating more synthetic samples.

{\noindent \bf  Self-filtering for better synthetic data.} Synthetic samples often suffer from degraded quality especially when running at large scale. Synthetic systems often rely on heuristics or rule-based filtering techniques to filter out bad-quality samples. Because  \sys pipeline is closed-loop, it implicitly provides self-filtering functionality. To check the quality of the synthetic text, we can detect if the synthetic text ($1d$) contains the visual tags ($1e$), to filter out potentially misaligned or lower quality synthetic text samples, we define that at least some ratio $p_{f}$ of the visual tags exist. For the synthetic image, we run it through the VTM again and output the visual tags, then we do the same check against the starting node text ($1b$ or $1d$). Later in \cref{fig:filtering}, we will show that self-filtering improves the synthetic samples.

\section{Experiments}

\subsection{Setup}

{\noindent \bf Tasks and Datasets.} We adopt the CLIP~\citep{radfordLearningTransferableVisual2021}  model architecture for multimodal representation learning. 
For pretraining CLIP models, we use two public image-text datasets: CC3M~\citep{sharmaConceptualCaptionsCleaned2018} and CC12M~\citep{changpinyoConceptual12MPushing2021}. %
To evaluate the representation quality of pretrained CLIP models, we measure the zero-shot performance on classification, retrieval, and compositional reasoning tasks. For image classification, we use 25 common vision datasets, including five ImageNet~\citep{dengImageNetLargescaleHierarchical2009,rechtImageNetClassifiersGeneralize2019} variants and the tasks from the VTAB benchmark~\citep{zhaiLargescaleStudyRepresentation2020}. We list the detailed dataset information in \cref{sec:dataset-details}. We use COCO~\citep{linMicrosoftCOCOCommon2014} and Flickr30k~\citep{plummerFlickr30kEntitiesCollecting2015} for image-to-text and text-to-image retrieval tasks and report the metrics in recall@1. SugarCrepe~\citep{hsiehSugarCrepeFixingHackable2023} is a recent benchmark that measures the compositional understanding of vision-language models, we report the zero-shot accuracy numbers. %
Additionally, to study the effects of \sys on long-tail tasks, we evaluate the task accuracy of Places-LT and ImageNet-LT datasets~\citep{liuLargeScaleLongTailedRecognition2019a} by augmenting the tail classes with \sys synthetic data.

{\noindent \bf Training and Baselines.} Note that \sys itself does not require any training. We conduct pretraining experiments on CLIP models to evaluate the quality of synthetic data. We use ViT-B/16~\citep{dosovitskiyImageWorth16x162020} architecture for the CLIP vision backbone. For a fair comparison, we train all models for the same number of iterations on the original dataset (baseline) and the dataset mixed with \sys augmented samples.
 We use \sys-cap to denote the original image and synthetic text pair $(1a, 1d)$ from synthesis path $SP(1)$. \sys-img stands for the synthetic image and original text pair $(1b, 1c)$ from synthesis path $SP(4)$. %
\sys-capimg means the synthetic image and text pair  $(1d, 1c)$ from synthesis path $SP(3)$. We define \sys-mix as taking one image-text pair from \sys-cap and another from \sys-capimg. We do not take \sys-img image-text pairs since we found the original texts are noisy and thus a substantial portion of synthetic images are bad quality. We refer \sys-mix as the default setting unless otherwise stated. We list detailed information in~\cref{sec:training-details}.%

{\noindent \bf \sys Models.} For the VTM, we adopt a hybrid approach by default, we combine the tags from a captioning plus tag extraction pipeline and an advanced multi-label image classifier. We use a recent vision foundation model called Florence-large~\citep{xiaoFlorence2AdvancingUnified2024} to generate detailed image descriptions and then extract the objects, attributes, and relations using the Qwen2-7B-Instruct~\citep{yangQwen2TechnicalReport2024} LLM. 
Then we use an accurate image classifier, the huge variant of CatLIP~\citep{mehtaCatLIPCLIPlevelVisual2024}, to output multiple high-confidence objects and attributes. We show later in \cref{sec:ablation} that this hybrid VTM provides the best visual tags compared with using individual approach alone. For the LLM, we use Mistral-NeMo-instruct model~\citep{aiMistralNeMo2024} by default due to its strong instruction-following capability. We choose the stable-diffusion-xl-base-1.0~\citep{podell2024sdxl} for the text-to-image model by default. We describe the detailed setup in \cref{sec:inference-details}. In \cref{sec:ablation}, we study different pretrained models for each of the three modules in \sys. 

\subsection{Main Results}

\begin{table}[t!]
\begin{minipage}[t]{0.48\linewidth}

\centering
    \caption{\small{Comparison of the zero-shot classification accuracy between the baseline and \sys. We report top-1 accuracy for 20 commonly used downstream vision datasets, including  12 tasks in the VTAB benchmark~\citep{zhaiLargescaleStudyRepresentation2020} and 8 other ones. %
}}\vspace{-0.5em}
\scriptsize
\begin{tabular}{@{}l|rr|rr@{}}
\toprule
\multirow{2}{*}{\bf Data $\backslash$ Model}               & \multicolumn{2}{c|}{CC3M} & \multicolumn{2}{c}{CC12M} \\ 
                  & CLIP   & \sys  & CLIP   & \sys   \\ \midrule
CIFAR-10               & 41.5   & \bf 70.3            & 75.4   & \bf 82.6             \\
CIFAR-100              & 14.1   & \bf 34.5            & 47.5   & \bf 53.4             \\
CLEVR Counts           & 7.1    & \bf 11.7            & 15.2   & \bf 22.1             \\
CLEVR Distance         & 16.1   & \bf 19.8            & \bf 18.6   & 18.0             \\
Caltech-101            & 43.8   & \bf 68.0            & \bf 76.5   & 76.2             \\
Country211             & 0.4    & \bf 0.6             & 1.1    & \bf 1.3              \\
DTD                    & 11.6   & \bf 17.9            & 23.5   & \bf 29.1             \\
EuroSAT                & 12.5   & \bf 15.1            & 25.4   & \bf 27.2             \\
FGVC Aircraft          & \bf 1.3    & 0.8             & 0.7    & \bf 1.8              \\
Food-101               & 9.5    & \bf 23.1            & 53.4   & \bf 61.0             \\
GTSRB                  & 4.6    & \bf 9.7             & 14.5   & \bf 19.1             \\
KITTI                  & \bf 30.2   & 19.5            & 33.9   & 33.9             \\
Oxford Flowers         & 10.8   & \bf 24.8            & 34.5   & \bf 38.9             \\
Oxford-IIIT Pet        & 3.1    & \bf 7.9             & 8.0    & \bf 9.4              \\
PatchCamelyon          & \bf 50.0   & 48.6            & \bf 52.7   & 50.4             \\
RESISC45               & 17.7   & \bf 27.6            & 36.7   & \bf 39.5             \\
STL-10                 & 70.4   & \bf 90.4            & 92.8   & \bf 94.0             \\
SUN397                 & 30.7   & \bf 44.3            & 54.1   & \bf 58.1             \\
SVHN                   & \bf 12.2   & 6.8             & 10.6   & \bf 14.0             \\
Stanford Cars          & 0.6    & 0.6             & \bf 2.3    & 2.0              \\ \midrule
Average                & 19.4   & \bf 27.1 \textcolor{green}{(+7.7)}           & 33.9   & \bf 36.6    \textcolor{green}{(+2.5)}          \\ \bottomrule
\end{tabular}
\label{tab:zs-common}
\end{minipage}
\hfill
\begin{minipage}[t]{0.48\linewidth}
\centering
    \caption{\small{Zero-shot top-1 accuracy between the baseline and \sys on 6 ImageNet datasets. %
}}\vspace{-0.5em}
\scriptsize
\label{tab:zs-imagenet}
\begin{tabular}{@{}l|rr|rr@{}}
\toprule
\multirow{2}{*}{\bf Data $\backslash$ Model}        & \multicolumn{2}{c|}{CC3M} & \multicolumn{2}{c}{CC12M} \\ 
           & CLIP     & \sys     & CLIP      & \sys     \\ \midrule
ImageNet-1K     & 20.2     & \bf 25.3          & 39.6      & \bf 41.2          \\
ImageNet-V2     & 11.0     & \bf 20.7          & 34.0      & \bf 35.5          \\
ImageNet-S      & 3.5      & \bf 12.4          & 28.3      & \bf 33.8          \\
ImageNet-A      & 3.0      & \bf 6.5           & 12.0      & \bf 14.9          \\
ImageNet-O      & 18.6     & \bf 30.7          & 44.2      & \bf 45.9          \\
ImageNet-R      & 11.6     & \bf 28.4          & 47.6      & \bf 55.1          \\ \midrule

Average      & 11.3     & \bf 20.7 \textcolor{green}{(+9.4)}         & 34.3      & \bf37.7  \textcolor{green}{(+3.4)}        \\ \bottomrule
\end{tabular}
\vspace{1.5em}
\centering
    \caption{\small{Zero-shot retrieval evaluation on the Flickr and COCO datasets. We report the recall@1 numbers. {I2T} means image-to-text retrieval, and {T2I} denotes text-to-image retrieval. %
}}\vspace{-0.5em}
\scriptsize
\label{tab:zs-rtr}
\begin{tabular}{@{}l|rr|rr@{}}
	\toprule
	{\multirow{2}{*}{\bf Data $\backslash$ Model}} & \multicolumn{2}{c|}{CC3M} & \multicolumn{2}{c}{CC12M} \\
                   & CLIP & \sys                                & CLIP & \sys                               \\ \midrule
	COCO I2T   & 10.9 & \bf 32.3                                & 40.5 & \bf 49.8                               \\
	COCO T2I   & 7.6  & \bf 19.8                                & 26.7 & \bf 32.2                               \\ \midrule
	Flickr I2T & 21.3 & \bf 57.3                                & 65.5 & \bf 77.2                               \\
	Flickr T2I & 14.8 & \bf 39.0                                & 48.9 & \bf 58.2                               \\ \midrule
	{Average}  & 13.7 & \bf 37.1 \textcolor{green}{(+23.4)} & 45.4 & \bf 54.4 \textcolor{green}{(+9.0)} \\ \bottomrule
\end{tabular}
\end{minipage}
\end{table}

{\noindent \bf Image Classification Evaluation.} We conduct the zero-shot evaluation for image classification tasks. \cref{tab:zs-common} shows the results across 20 commonly used vision datasets and \cref{tab:zs-imagenet} shows the results of 6 ImageNet-related datasets. Notably, \sys outperforms the baseline consistently by 2.5\% to 9.4\% for the CLIP models trained on the CC3M and CC12M datasets. We observe that \sys significantly improves the zero-shot performance (by over 7.7\%) by augmenting smaller datasets like CC3M, while the performance gains become smaller on larger datasets like CC12M.

{\noindent \bf Image-Text Retrieval Evaluation.} We evaluate the zero-shot image-text retrieval performance for our \sys and baseline CLIP models and present the recall@1 results in \cref{tab:zs-rtr}.  \sys substantially improves the text-to-image and image-to-text retrieval recall by up to 24\% and 36\% for the Flickr dataset, and overall improves recall by 23.4\% on average for CC3M models. \sys also brings over 9\% retrieval gains for CC12M models on average. The improvements show that data samples from \sys have better coverage of visual concepts. %

{\noindent \bf Compositional Reasoning Results.} A key strength in \sys is the inclusion of visual tags that contain objects, attributes and relations from an image. To understand how the fine-grained visual attributes and relations affect visual reasoning performance, we evaluate \sys and baseline on the SugarCrepe~\cite{hsiehSugarCrepeFixingHackable2023} benchmark which measures the compositional reasoning capability of vision language models. We present the results in \cref{tab:sg}. \sys improves the baseline CLIP compositional reasoning by a large margin (4.5\% for CC3M and 3\% for CC12M on average). Note that most of the improvements come from the attribute and relation forms in the \textsc{Replace} and \textsc{Swap} categories, for example, \sys on CC3M improves the \textsc{Replace} relation accuracy by 4.3\% and \textsc{Swap} attribute by 14.8\%, indicating \sys models are robust to the attribute and relation changes. %

\begin{table*}[t!]
  \centering
  \caption{\small We evaluate the compositional reasoning accuracy on the SugarCrepe~\citep{hsiehSugarCrepeFixingHackable2023} benchmark.}
  \vspace{-.5em}
  \scriptsize
    \begin{tabular}{cccccccccl} 
    \toprule
 \multirow{2}{*}{\textbf{Data}}  & \multirow{2}{*}{\textbf{Model}} & \multicolumn{2}{c}{\bf \textsc{Add}} & \multicolumn{3}{c}{\bf \textsc{Replace}} &  \multicolumn{2}{c}{\bf \textsc{Swap}} & \multirow{2}{*}{\bf \textsc{Average}} \\ 
     \cmidrule(lr){3-4}\cmidrule(lr){5-7}\cmidrule(lr){8-9}
   &  & Attribute  & Object & Attribute & Object & Relation & Attribute & Object & \\ 
    \midrule

 \multirow{2}{*}{CC3M} & {CLIP} & \bf  69.2 & 71.0 & 69.3 & 80.3 & 55.2 & 52.6 & 50.6 & 64.0 \\
    & {\sys} & 66.2 & 71.0 & \bf 73.1 & \bf  82.8 & \bf  59.5 & \bf  67.4 & \bf  59.6 & \bf 68.5 (\textcolor{green}{+4.5}) \\ \midrule
  \multirow{2}{*}{CC12M} &{CLIP} & 70.7 & 77.8 & 78.7 & 88.4 & 66.7 & 61.7 & 62.0 & 72.3 \\
   & {\sys} & \bf 71.7 & \bf 78.7 & \bf 82.6 & 88.3 & \bf 69.3 & \bf {72.7} & \bf 63.7 & \bf 75.3 ({\textcolor{green}{+3.0}}) \\
\bottomrule 
\end{tabular}
\label{tab:sg}
\vspace{-.5em}
\end{table*}

{\noindent \bf Comparison with Prior Work.} \sys pipeline is flexible and supports synthesizing data from different paths. Previous work like VeCLIP~\citep{laiVeCLIPImprovingCLIP2024} and LaCLIP~\citep{fanImprovingCLIPTraining2023} synthesizing new texts for the images by improving the captions. Though it is impossible to have a completely fair comparison with them\footnote{Factors that prohibit apple-to-apple comparison include training software, variations of CC3M samples due to missing images, exact hardware set up, etc.}, the synthetic texts from the synthesis path (2) in \sys provide similar effects. We present the results on CLIP ViT/B16 models trained on CC3M for the tasks reported in each work. \cref{tab:zs-velip} shows that \sys outperforms VeCLIP on most VTAB datasets and improves zero-shot accuracy by 4.8\% on average. \sys also surpasses VeCLIP by 7.9\% on the ImageNet 1K dataset. We observe a similar trend when comparing \sys with LaCLIP in \cref{tab:zs-laclip}. Specifically, \sys achieves an average of 3.4\% better accuracy than LaCLIP on 15 common datasets and 2.3\% better accuracy on ImageNet 1K.

\begin{table}[t!]
    \centering
    \caption{\small{Comparison of the zero-shot classification accuracy between VeCLIP~\citep{vasuCLIPQualityCaptions2024} and \sys for CLIP trained on the CC3M. We report top-1 accuracy (\%) for the VTAB benchmark~\citep{zhaiLargescaleStudyRepresentation2020} across 9 tasks (6 from natural and 3 from specialized sets). We highlight the best numbers in {\bf{bold}}. 
}}
    \vspace{-0.5em}
    \resizebox{0.98\textwidth}{!}{
    \begin{tabular}{c|cccccc|ccc|c|c}
    \toprule[1.2pt]
     \multirow{2}{*}{\bf Model}  & \multicolumn{6}{c|}{ \bf Natural Sets}  & \multicolumn{3}{c|}{ \bf Specialized Sets } & \multirow{2}{*}{\bf Average } & \multirow{2}{*}{\bf ImageNet 1K }\\
    &  Caltech101 & CIFAR100 & SVHN & DTD & OxPet & Flowers102 & EuroSAT & RESISC45 & Camelyon &  \\
    
    \midrule 
    \small CLIP & 39.50 & 9.83 &  \bf 20.89 & 7.42 & 7.44 & 10.40 &  11.94 & 7.93 & 50.65 & 18.45 & 5.46 \\
    \small VeCLIP &  54.30 &  17.74 &  18.74 &  11.23 & \bf 10.09 & \bf 22.75 &  7.35 &  16.54 & \bf 52.52 &  23.48 & 15.98 \\
     \small \sys & \bf 66.10 & \bf 34.09 & 17.66 & \bf 16.76 & 7.77 & 15.55 & \bf 20.83 & \bf 24.59 & 50.79 & \bf 28.24 & \bf 23.82 \\
    \bottomrule[1.2pt]
    \end{tabular}\label{tab:zs-velip}
    }\vspace{-.5em}
\end{table}

\begin{table}[t!]
\vspace{-0.5em}
\centering
\caption{\small{We report the zero-shot performance on ImageNet 1K and 15 common downstream datasets for both LaCLIP~\citep{fanImprovingCLIPTraining2023} and \sys for CLIP trained on CC3M. We highlight the best numbers in {\bf{bold}}. 
}}\vspace{-0.5em}

\resizebox{.98\textwidth}{!}{\centering
\begin{tabular}{c@{\hspace{0.5em}}|c@{\hspace{0.5em}}c@{\hspace{0.5em}}c@{\hspace{0.5em}}c@{\hspace{0.5em}}c@{\hspace{0.5em}}c@{\hspace{0.5em}}c@{\hspace{0.5em}}c@{\hspace{0.5em}}c@{\hspace{0.5em}}c@{\hspace{0.5em}}c@{\hspace{0.5em}}c@{\hspace{0.5em}}c@{\hspace{0.5em}}c@{\hspace{0.5em}}c@{\hspace{0.5em}}|c@{\hspace{0.5em}}|c@{\hspace{0.5em}}}
    \toprule%
    \small\bf Model&
    \rotatebox[origin=lb]{90}{\smash{\small Food-101}} & \rotatebox[origin=lb]{90}{\smash{\small CIFAR-10}} & \rotatebox[origin=lb]{90}{\smash{\small CIFAR-100}} & \rotatebox[origin=lb]{90}{\smash{\small SUN397}} &
    \rotatebox[origin=lb]{90}{\smash{\small Cars}} & \rotatebox[origin=lb]{90}{\smash{\small Aircraft}} & \rotatebox[origin=lb]{90}{\smash{\small DTD}} & \rotatebox[origin=lb]{90}{\smash{\small Pets}} & \rotatebox[origin=lb]{90}{\smash{\small Caltech-101}} &
    \rotatebox[origin=lb]{90}{\smash{\small Flowers}} & \rotatebox[origin=lb]{90}{\smash{\small STL-10}} & \rotatebox[origin=lb]{90}{\smash{\small EuroSAT}} &
    \rotatebox[origin=lb]{90}{\smash{\small RESISC45}} & \rotatebox[origin=lb]{90}{\smash{\small GTSRB}} & \rotatebox[origin=lb]{90}{\smash{\small Country211}}  & \rotatebox[origin=lb]{90}{\smash{\small \bf Average}} & \rotatebox[origin=lb]{90}{\smash{\small \bf ImageNet}} \\
    \midrule
    \small CLIP   & 10.3     & 54.9     & 21.8     & 25.0     & 0.8     & 1.4     & 10.5     & 12.8     & 43.3     & 10.2     & 77.6     & 14.1     & 19.1     & 6.9 & 0.6     & 20.6     & 15.8     \\
    \small LaCLIP & 14.2 & 57.1 & 27.5 & 35.1 & \bf 1.6 & \bf 1.6 & 16.6 & \bf 15.6 & 52.7 & 14.7 & 86.2 & 15.0 & 24.3 & 6.4     & \bf 1.0 & 24.6 & 21.5 \\
    \small \sys   & \bf 17.8 & \bf 69.5 & \bf 34.1 & \bf 44.9 & 0.7 & 1.2 & \bf 16.8 & 7.8 & \bf 66.1 & \bf 15.5 & \bf 88.3 & \bf 20.8 & \bf 24.6 & \bf 10.9     & 0.7 & \bf 28.0 & \bf 23.8 \\
    \bottomrule%
\end{tabular}\label{tab:zs-laclip}
}\vspace{-.8em}
\end{table}

\subsection{Performance on Long-tail Tasks.}
Real-world data often have long-tail distributions. Much recent research~\citep{shiLongTailLearningFoundation2024,liuLargeScaleLongTailedRecognition2019a} has focused on developing new learning methods for long-tail recognition tasks. Data augmentation remains an important solution, especially when the tail classes only have a few samples. In this section, we evaluate the effectiveness of synthetic samples from \sys for long-tail tasks. 

{\noindent \bf Setup.} We conduct experiments on the ImageNet-LT~\citep{liuLargeScaleLongTailedRecognition2019a} and Places-LT~\citep{liuLargeScaleLongTailedRecognition2019a} datasets. ImageNet-LT is a subset of the original ImageNet-2012~\citep{dengImageNetLargescaleHierarchical2009} and contains 115.8K images from 1000 classes, with 5 to 1280 images per class. Places-LT is even more imbalanced and contains 62.5K images from 365 classes, with 5 to 4980 images per class. The test sets of both datasets are balanced. Following the same setup in~\citep{liuLargeScaleLongTailedRecognition2019a}, we report the overall accuracy as well as the accuracy across the head ($>$100 images), medium (20$\sim$100), and tail ($<$20) classes. We take the same baseline in \citep{shiLongTailLearningFoundation2024} and fine-tune the classifier head of a pretrained CLIP model (ViT-B/16) for 10 epochs (or the same number of iterations for \sys). For \sys synthetic samples, we choose the synthetic path $SP(3)$ to generate synthetic images for the tail classes. We mix the \sys image samples with the original training set of each dataset. We describe more details in \cref{sec:dataset-details}. %

{\noindent \bf Key Results.} \cref{tab:longtail} shows that \sys improves the tail class accuracy by 21.3\% on ImageNet-LT and by 16.2\% on Places-LT. Synthetic samples from \sys also improve the overall and medium class accuracy by 3$\sim$6\%, though slightly decrease the head class accuracy. %

\begin{table}[t!]
\centering
\caption{\small{Long-tail accuracy on the ImagetNet-LT and Places-LT datasets for the baseline and \sys models.}} 
\vspace{-.5em}
\resizebox{.98\textwidth}{!}{\centering
\begin{tabular}{@{}c|llll|llll@{}}
\toprule
\multirow{2}{*}{Model} & \multicolumn{4}{c|}{\bf ImageNet-LT} & \multicolumn{4}{c}{\bf Places-LT}  \\
& Overall &  Tail &  Medium &  Head &  Overall &  Tail &  Medium &  Head \\ \midrule
Baseline & 60.8        & 13.8     & 56.7       & \bf 82.6 & 34.9        & 8.2      & 31.3       & \bf 53.7 \\
\sys     & \bf 66.2 (\textcolor{green}{+5.4})   & \bf 35.1 (\textcolor{green}{+21.3}) & \bf 62.8 (\textcolor{green}{+6.1})   & 81.4     & \bf 38.6 (\textcolor{green}{+3.7})   & \bf 24.4 (\textcolor{green}{+16.2}) & \bf 34.6 (\textcolor{green}{+3.3})   & 51.2     \\ \bottomrule
\end{tabular}\label{tab:longtail}
}
\vspace{-.5em}
\end{table}

\subsection{Analysis}
\label{sec:analysis}

\mdfdefinestyle{mdf2}{innertopmargin=0.2em,innerbottommargin=0.2em,%
innerleftmargin=0.2em,innerrightmargin=0.2em,%
roundcorner=2pt,hidealllines=true}

\begin{wrapfigure}{r}{0.45\textwidth}
\vspace{-4.5em}
\begin{mdframed}[style=mdf2]
\vspace{-0.5em}
\includegraphics[width=\linewidth]{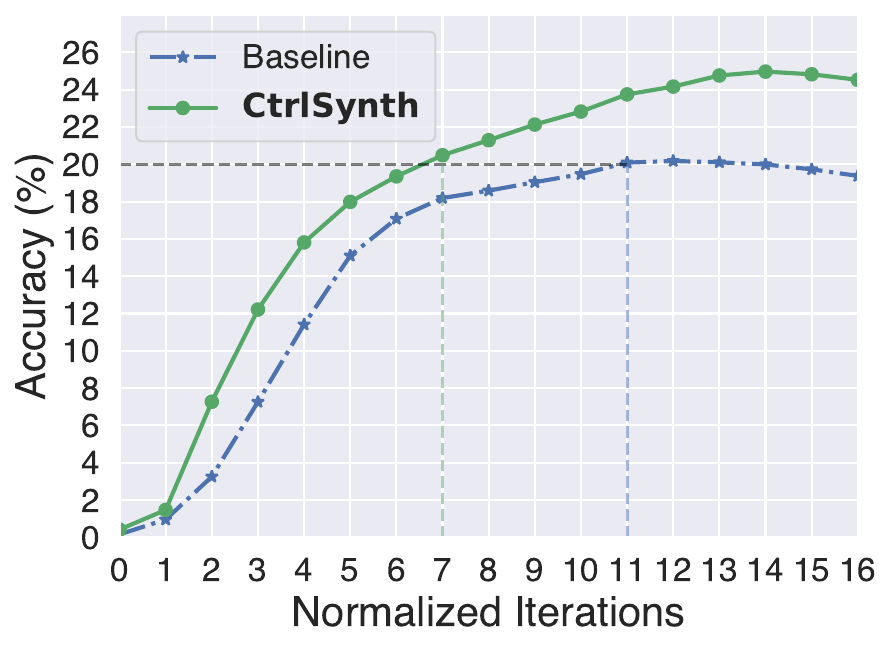}
\vspace{-2em}
\caption{\small{Data efficiency comparison between baseline and \sys for pretraining CLIP models on CC3M. We normalize the iterations by dividing the total iterations with checkpoint steps.}}%
\end{mdframed}
\label{fig:data-efficiency}
\vspace{-3em}
\end{wrapfigure}

\paragraph{Data-Efficiency of \sys in Training CLIP.} To study the data efficiency of \sys samples, we plot the top1 zero-shot accuracy of the ImageNet validation set in \cref{fig:data-efficiency} for the baseline and \sys CLIP models trained on CC3M. \sys reaches the 20\% accuracy with 40\% fewer iterations than the baseline, indicating that using \sys samples is more data-efficient.  %

\begin{figure*}[htb!]
    \centering
  \rotatebox[origin=lb]{90}{\smash{\small Original}} \hfill
  \subcaptionbox*{\tiny wisteria decorates a black and white timbered cottage}%
  {\includegraphics[width=0.24\linewidth]{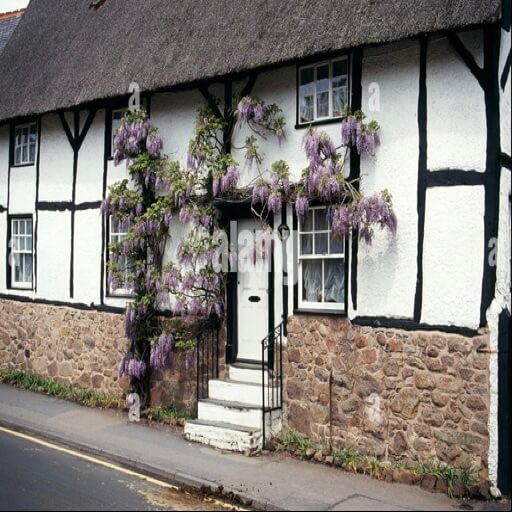}\vspace{-0.5em}}
  \hfill
  \subcaptionbox*{\tiny fresh red shrimps for sale at a market}%
  {\includegraphics[width=0.24\linewidth]{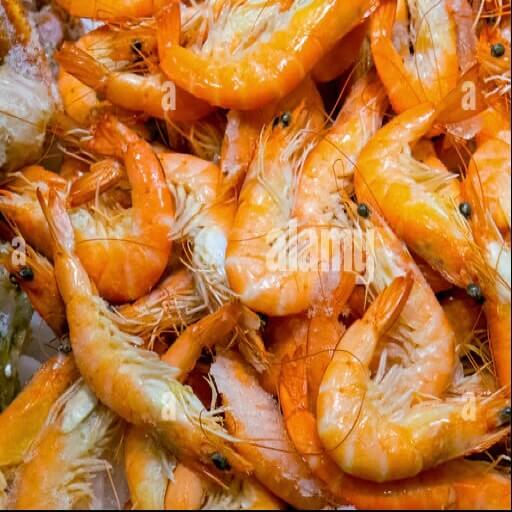}\vspace{-0.5em}}
  \hfill
  \subcaptionbox*{\tiny view into the living room .}%
  {\includegraphics[width=0.24\linewidth]{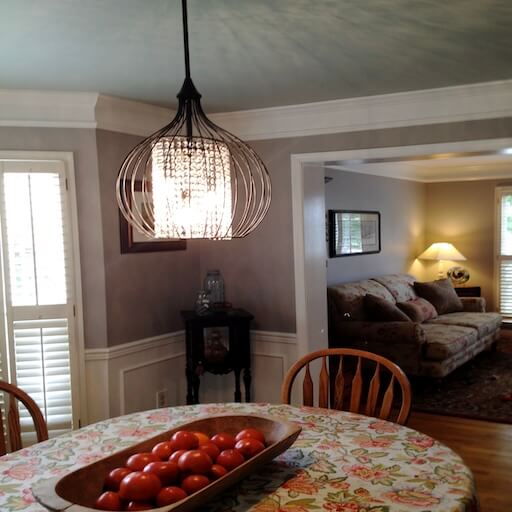}\vspace{-0.5em}}
  \hfill
  \subcaptionbox*{\tiny tuscan sun in the landscape}%
  {\includegraphics[width=0.24\linewidth]{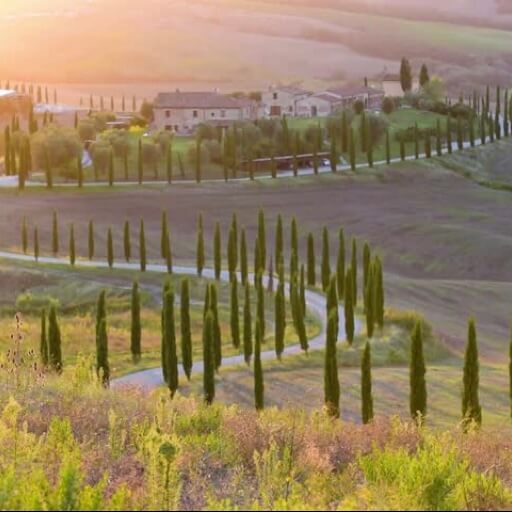}\vspace{-0.5em}}
 \hspace{\fill}\\
 \rotatebox[origin=lb]{90}{\smash{\small Synthetic}} \hfill
 \subcaptionbox*{\tiny wisteria, with its purple flowers, hangs from the eaves and twines around the wrought iron railing, decorating the small porch of the old, black and white painted cottage.}%
  {\includegraphics[width=0.24\linewidth]{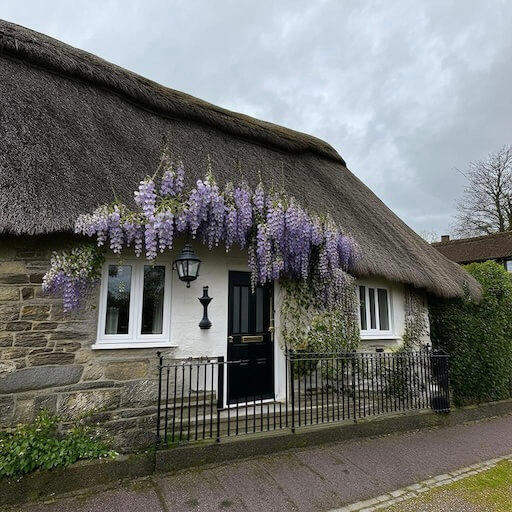}\vspace{-0.5em}}
  \hfill
  \subcaptionbox*{\tiny freshly caught, red shrimp, arranged in a pile at the bustling seafood market, their small black spots visible, overlapping and cooked to perfection, surrounded by a blur of herbs and spices.}%
  {\includegraphics[width=0.24\linewidth]{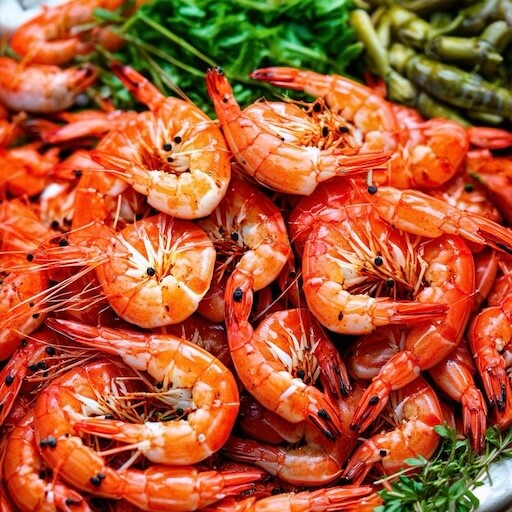}\vspace{-0.5em}}
  \hfill
  \subcaptionbox*{\tiny a living room, where a large pendant light hangs from the ceiling. on the right, a wooden table is covered with a floral tablecloth, set with a wooden bowl of red tomatoes and a lamp.}%
  {\includegraphics[width=0.24\linewidth]{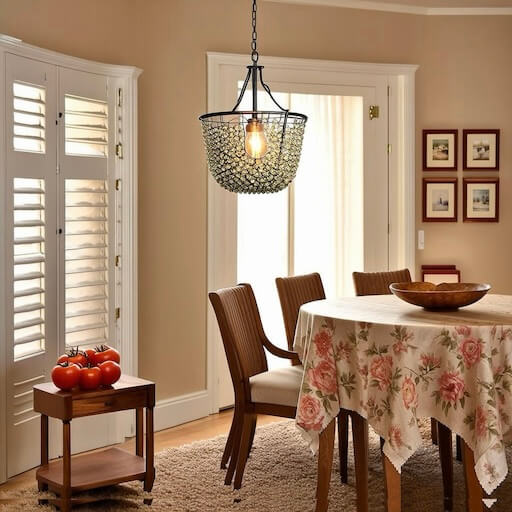}\vspace{-0.5em}}
  \hfill
  \subcaptionbox*{\tiny tuscan sun casting a warm, orange glow over the serene italian countryside, with tall cypress trees arranged in neat rows along the winding road, the sun setting in the background.}%
  {\includegraphics[width=0.24\linewidth]{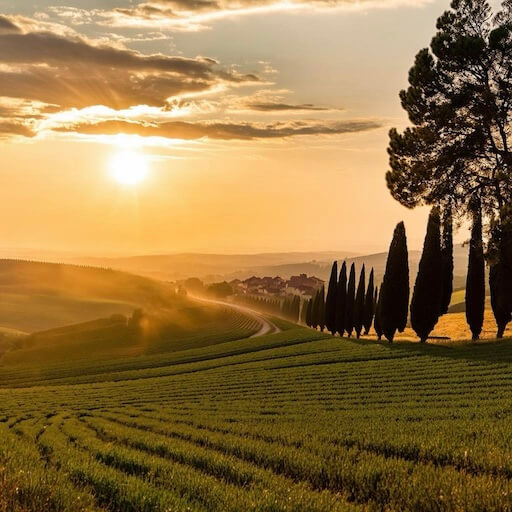}\vspace{-0.5em}}
\caption{Randomly selected CC3M examples of real images and captions (the first row) with their corresponding \sys synthetic samples (the second row).%
}
\label{fig:demo}
\vspace{-1.5em}
\end{figure*}

\paragraph{Statistics and visualization of \sys Samples.} In this section, we provide the statistics for the synthetic samples from \sys.  \cref{fig:demo} shows examples of \sys images and texts compared with the original real samples. We observe that the text samples from \sys are usually longer and contain richer information about the image. On average, \sys texts have over 60 words while original captions contain 8 words. We plot the histogram of the number of words in \cref{fig:words-freq} at \cref{sec:analysis-details}.

\begin{figure}[t!]
    \centering
 \subcaptionbox{\small Accuracy versus filtering thresholds.}%
  {\includegraphics[width=0.45\linewidth]{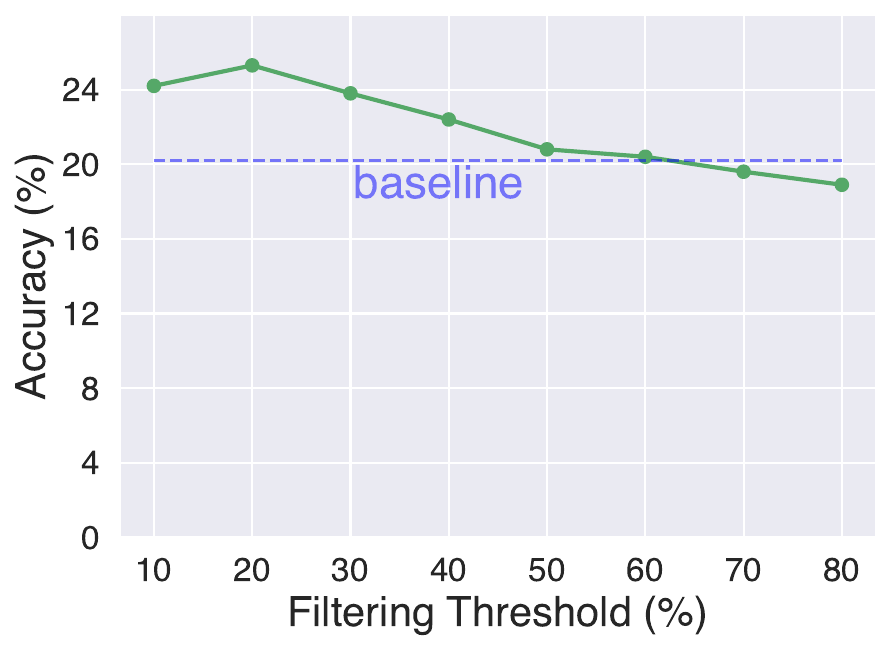}\vspace{-0.5em}\label{fig:filtering}}
  \hspace{\fill}
  \subcaptionbox{\small Accuracy versus mixing ratios.}%
  {\includegraphics[width=0.45\linewidth]{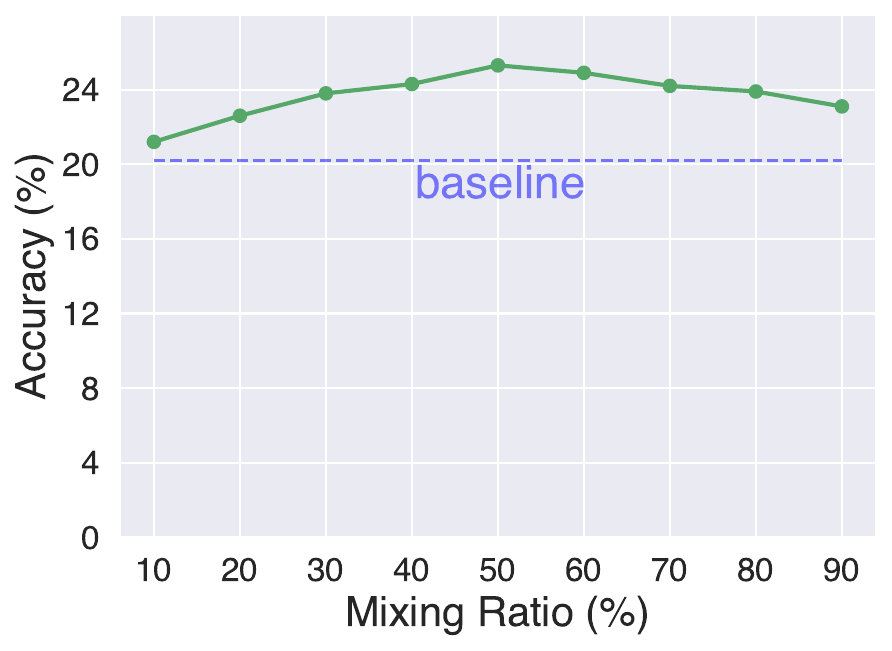}\vspace{-0.5em}\label{fig:mixing}}
\vspace{-0.5em}
\caption{Study of different filtering thresholds and mixing ratios of \sys samples. The accuracy numbers are top1 zero-shot accuracy on the ImageNet-1K validation set. The CLIP models are trained on the CC3M dataset and \sys samples. }
\vspace{-1.2em}
\end{figure}

\paragraph{Effects of Self-Filtering.} \sys provides off-the-shelf self-filtering to control the quality of synthetic samples. We study the effects of applying different filtering thresholds $p_{f}$ for the synthetic text and image. %
We set the same filtering thresholds for both synthetic text and image samples. Intuitively, a higher threshold filters out more synthetic samples thus providing better quality samples that align with original real samples. On the contrary, a lower threshold keeps relatively less aligned samples but encourages more diverse samples. \cref{fig:filtering} plots the zero-shot accuracy numbers of CLIP model on ImageNet under different threshold settings, we show that thresholds 10\%$\sim$30\% provide similar accuracy numbers and setting the filtering threshold to 20\% provides the best accuracy. Thresholds higher than 50\% do not provide accuracy gains, likely because the aligned synthetic samples lack diversity and fail to augment the original samples.

\paragraph{Mixing Ratios of Synthetic Samples.} To better understand how the synthetic image text samples improve CLIP model training, we study different ratios ($p_r$) of mixing \sys samples with original real ones. During CLIP training, we randomly sample the original sample with probability $0<p_r<1$ and our sample with $1-p_r$. \cref{fig:mixing} shows that even adding a small portion ($<20$\%) of \sys samples improves the zero-shot accuracy while mixing with 50\% provides best accuracy gains. Further higher mixing ratios show diminishing improvements though still better than the baseline that uses all real data. %

\subsection{Ablation Study}
\label{sec:ablation}
In this section, we evaluate the effectiveness of visual tags, the impact of using different pretrained models in the \sys pipeline, and mixing and filtering effects for \sys samples. We use the same text and image control policy described in \cref{sec:synth} for all settings. We experiment with CC3M dataset for CLIP pretraining and report the accuracy on the SugarCrepe benchmark, zero-shot accuracy of common downstream vision tasks (same tasks in \cref{tab:zs-common}), and top1 accuracy on the ImageNet 1k validation set. 

\begin{table}[t!]
\centering
\caption{\small{Evaluation of using different models, visual tags, and synthetic samples in \sys. '-' denotes the same value from the last row (default setting).}} 
\resizebox{.98\textwidth}{!}{\centering
\begin{tabular}{@{}llllccc@{}}
\toprule
Study                              & Model              & Tags         & Samples              & Common Tasks & ImageNet-1K & SugarCrepe \\ \midrule
\multirow{3}{*}{Models}  & Qwen2-7B, SDXL     & - & - & 24.7         & 23.5        & 65.1       \\
                                   & Qwen2-7B, SD3M     & - & - & 26.1         & 23.8        & 65.2       \\
                                   & Mistral-Nemo, SD3M & - & - & 26.6         & 25.1        & 68.1       \\ \midrule
\multirow{2}{*}{Tags}       & - & Obj          & - & 26.4         & 24.7        & 64.3       \\
                                   & - & Obj+Attr     & - & 26.2         & 24.8        & 65.4       \\ \midrule
\multirow{3}{*}{Samples} & - & - & \sys-cap, SP(1)       & 26.2         & 24.5        & 67.2       \\
                                   & - & - & \sys-img, SP(4)       & 22.1         & 21.8        & 64.4       \\
                                   & - & - & \sys-capimg, SP(3)    & 26.5         & 24.8        & 67.5       \\ \midrule
\sys                                & Mistral-Nemo, SDXL & Obj+Attr+Rel & \sys-mix & 27.1         & 25.3        & 68.5       \\ 
                                   \bottomrule %
\end{tabular}\label{tab:ablation}
}
\end{table}

{\noindent \bf Different Pretrained Models.} We choose an alternate LLM and a different text-to-image model to understand how different pretrained models affect the quality of synthetic samples. \sys pipeline is flexible so we can easily swap the pretrained LLM and text-to-image models. Specifically, we use Qwen2-7B~\citep{yangQwen2TechnicalReport2024} for the LLM and Stable Diffusion 3 Medium~\citep{esserScalingRectifiedFlow2024} (SD3M) for the text-to-image model. Comparing the first and last rows in \cref{tab:ablation}, we find using a smaller LLM like Qwen2-7B degrades the task performance on all three tasks, indicating that using a strong LLM is key to synthesizing high quality texts. The accuracy boost (+3\%) on SugarCrepe benchmark shows the LLM is effective in recombining the visual tags in a compositional way to form diverse synthetic texts. 
We also point out that using a more recent diffusion model like SD3M provides similar task performance numbers, this is likely because SD3M has fewer (2B versus 3.5B) parameters compared to SDXL, limiting the image generation capability.

{\noindent \bf Effectiveness of Visual Tags.} We study the effects of using different categories of visual tags, \ie, using only objects (Obj), objects plus attributes (Obj+Attr), and all categories including relations (Obj+Attr+Rel). In \cref{tab:ablation}, comparing the second and last row, we show attributes marginally improve the CLIP performance on compositional reasoning but not much on zero-shot vision tasks. Importantly, visual relations improves the performance  on all three tasks, and significantly improves compositional reasoning performance by over 4\%.

{\noindent \bf \sys Samples from Different Synthesis Paths.} \sys pipeline supports synthesizing images or texts from different paths, we evaluate their quality by measuring the downstream task accuracy of the CLIP models trained on them. The penultimate and last rows in \cref{sec:ablation} show all \sys samples provides performance gains on downstream tasks, except the \sys-img samples where they do not improve compositional reasoning performance. \sys-img samples have the least augmentation benefits and are likely due to the original real texts are noisy and thus the generated images are not of high quality. Notably, mixing with synthetic captions (\sys-cap, \sys-capimg, and \sys-mix) provides meaningful augmentation benefits, highlight the importance of using LLMs to recombine the visual tags.

\section{Conclusion}
Synthetic data emerges as a viable solution to address challenges in curating high-quality samples from noisy, misaligned, and long-tail web data. However, existing data synthesis pipelines are rigid and the generation process is hard to control and thus being tailored for ad hoc use cases. 
We develop \sys, a new image-text synthesis pipeline that allows users to control the data generation in a fine-grained way. \sys decomposes the semantics of images and texts into basic elements and uses pretrained foundation models to recompose them based on specified control policies. This way, \sys provides flexible and diverse image-text samples. Synthetic samples from \sys improve the long-tail task performance by a large margin. They also significantly boost the zero-shot and compositional capability of CLIP models and enable data-efficient multimodal learning.

\bibliography{related}

\begin{thebibliography}{65}
\providecommand{\natexlab}[1]{#1}
\providecommand{\url}[1]{\texttt{#1}}
\expandafter\ifx\csname urlstyle\endcsname\relax
  \providecommand{\doi}[1]{doi: #1}\else
  \providecommand{\doi}{doi: \begingroup \urlstyle{rm}\Url}\fi

\bibitem[AI(2024)]{aiMistralNeMo2024}
Mistral AI.
\newblock Mistral {NeMo}, July 2024.
\newblock URL \url{https://mistral.ai/news/mistral-nemo/}.

\bibitem[Bossard et~al.(2014)Bossard, Guillaumin, and Van~Gool]{bossardFood101MiningDiscriminative2014}
Lukas Bossard, Matthieu Guillaumin, and Luc Van~Gool.
\newblock Food-101 – {Mining} {Discriminative} {Components} with {Random} {Forests}.
\newblock In David Fleet, Tomas Pajdla, Bernt Schiele, and Tinne Tuytelaars (eds.), \emph{Computer {Vision} – {ECCV} 2014}, pp.\  446--461, Cham, 2014. Springer International Publishing.
\newblock ISBN 978-3-319-10599-4.
\newblock \doi{10.1007/978-3-319-10599-4_29}.

\bibitem[Changpinyo et~al.(2021)Changpinyo, Sharma, Ding, and Soricut]{changpinyoConceptual12MPushing2021}
Soravit Changpinyo, Piyush Sharma, Nan Ding, and Radu Soricut.
\newblock Conceptual {12M}: {Pushing} {Web}-{Scale} {Image}-{Text} {Pre}-{Training} {To} {Recognize} {Long}-{Tail} {Visual} {Concepts}.
\newblock In \emph{Proceedings of the {IEEE}/{CVF} {Conference} on {Computer} {Vision} and {Pattern} {Recognition} ({CVPR})}, pp.\  3558--3568, 2021.
\newblock URL \url{https://openaccess.thecvf.com/content/CVPR2021/html/Changpinyo_Conceptual_12M_Pushing_Web-Scale_Image-Text_Pre-Training_To_Recognize_Long-Tail_Visual_CVPR_2021_paper.html}.

\bibitem[Cheng et~al.(2017)Cheng, Han, and Lu]{chengRemoteSensingImage2017}
Gong Cheng, Junwei Han, and Xiaoqiang Lu.
\newblock Remote {Sensing} {Image} {Scene} {Classification}: {Benchmark} and {State} of the {Art}.
\newblock \emph{Proceedings of the IEEE}, 105\penalty0 (10):\penalty0 1865--1883, October 2017.
\newblock ISSN 1558-2256.
\newblock \doi{10.1109/JPROC.2017.2675998}.
\newblock URL \url{https://ieeexplore.ieee.org/document/7891544}.

\bibitem[Cimpoi et~al.(2014)Cimpoi, Maji, Kokkinos, Mohamed, and Vedaldi]{cimpoiDescribingTexturesWild2014}
Mircea Cimpoi, Subhransu Maji, Iasonas Kokkinos, Sammy Mohamed, and Andrea Vedaldi.
\newblock Describing {Textures} in the {Wild}.
\newblock In \emph{2014 {IEEE} {Conference} on {Computer} {Vision} and {Pattern} {Recognition}}, pp.\  3606--3613, June 2014.
\newblock \doi{10.1109/CVPR.2014.461}.
\newblock URL \url{https://ieeexplore.ieee.org/document/6909856}.

\bibitem[Coates et~al.(2011)Coates, Ng, and Lee]{coatesAnalysisSingleLayerNetworks2011}
Adam Coates, Andrew Ng, and Honglak Lee.
\newblock An {Analysis} of {Single}-{Layer} {Networks} in {Unsupervised} {Feature} {Learning}.
\newblock In \emph{Proceedings of the {Fourteenth} {International} {Conference} on {Artificial} {Intelligence} and {Statistics}}, pp.\  215--223. JMLR Workshop and Conference Proceedings, June 2011.
\newblock URL \url{https://proceedings.mlr.press/v15/coates11a.html}.

\bibitem[Deng et~al.(2009)Deng, Dong, Socher, Li, Li, and Fei-Fei]{dengImageNetLargescaleHierarchical2009}
Jia Deng, Wei Dong, Richard Socher, Li-Jia Li, Kai Li, and Li~Fei-Fei.
\newblock {ImageNet}: {A} large-scale hierarchical image database.
\newblock In \emph{2009 {IEEE} {Conference} on {Computer} {Vision} and {Pattern} {Recognition}}, pp.\  248--255. IEEE Computer Society, June 2009.
\newblock ISBN 978-1-4244-3992-8.
\newblock \doi{10.1109/CVPR.2009.5206848}.
\newblock URL \url{https://www.computer.org/csdl/proceedings-article/cvpr/2009/05206848/12OmNxWcH55}.

\bibitem[Dosovitskiy et~al.(2020)Dosovitskiy, Beyer, Kolesnikov, Weissenborn, Zhai, Unterthiner, Dehghani, Minderer, Heigold, Gelly, Uszkoreit, and Houlsby]{dosovitskiyImageWorth16x162020}
Alexey Dosovitskiy, Lucas Beyer, Alexander Kolesnikov, Dirk Weissenborn, Xiaohua Zhai, Thomas Unterthiner, Mostafa Dehghani, Matthias Minderer, Georg Heigold, Sylvain Gelly, Jakob Uszkoreit, and Neil Houlsby.
\newblock An {Image} is {Worth} 16x16 {Words}: {Transformers} for {Image} {Recognition} at {Scale}.
\newblock In \emph{International {Conference} on {Learning} {Representations}}, September 2020.
\newblock URL \url{https://openreview.net/forum?id=YicbFdNTTy}.

\bibitem[Dunlap et~al.(2023)Dunlap, Umino, Zhang, Yang, Gonzalez, and Darrell]{dunlapDiversifyYourVision2023}
Lisa Dunlap, Alyssa Umino, Han Zhang, Jiezhi Yang, Joseph~E. Gonzalez, and Trevor Darrell.
\newblock Diversify {Your} {Vision} {Datasets} with {Automatic} {Diffusion}-based {Augmentation}.
\newblock In \emph{Thirty-seventh {Conference} on {Neural} {Information} {Processing} {Systems}}, November 2023.
\newblock URL \url{https://openreview.net/forum?id=9wrYfqdrwk}.

\bibitem[Esser et~al.(2024)Esser, Kulal, Blattmann, Entezari, Müller, Saini, Levi, Lorenz, Sauer, Boesel, Podell, Dockhorn, English, Lacey, Goodwin, Marek, and Rombach]{esserScalingRectifiedFlow2024}
Patrick Esser, Sumith Kulal, Andreas Blattmann, Rahim Entezari, Jonas Müller, Harry Saini, Yam Levi, Dominik Lorenz, Axel Sauer, Frederic Boesel, Dustin Podell, Tim Dockhorn, Zion English, Kyle Lacey, Alex Goodwin, Yannik Marek, and Robin Rombach.
\newblock Scaling {Rectified} {Flow} {Transformers} for {High}-{Resolution} {Image} {Synthesis}, March 2024.
\newblock URL \url{http://arxiv.org/abs/2403.03206}.

\bibitem[Fan et~al.(2023)Fan, Krishnan, Isola, Katabi, and Tian]{fanImprovingCLIPTraining2023}
Lijie Fan, Dilip Krishnan, Phillip Isola, Dina Katabi, and Yonglong Tian.
\newblock Improving {CLIP} {Training} with {Language} {Rewrites}, October 2023.
\newblock URL \url{http://arxiv.org/abs/2305.20088}.

\bibitem[Fei-Fei et~al.(2006)Fei-Fei, Fergus, and Perona]{fei-feiOneshotLearningObject2006}
Li~Fei-Fei, R.~Fergus, and P.~Perona.
\newblock One-shot learning of object categories.
\newblock \emph{IEEE Transactions on Pattern Analysis and Machine Intelligence}, 28\penalty0 (4):\penalty0 594--611, April 2006.
\newblock ISSN 1939-3539.
\newblock \doi{10.1109/TPAMI.2006.79}.
\newblock URL \url{https://ieeexplore.ieee.org/document/1597116}.

\bibitem[Geiger et~al.(2013)Geiger, Lenz, Stiller, and Urtasun]{geigerVisionMeetsRobotics2013}
A~Geiger, P~Lenz, C~Stiller, and R~Urtasun.
\newblock Vision meets robotics: {The} {KITTI} dataset.
\newblock \emph{The International Journal of Robotics Research}, 32\penalty0 (11):\penalty0 1231--1237, September 2013.
\newblock ISSN 0278-3649.
\newblock \doi{10.1177/0278364913491297}.
\newblock URL \url{https://doi.org/10.1177/0278364913491297}.

\bibitem[Helber et~al.(2018)Helber, Bischke, Dengel, and Borth]{helberIntroducingEurosatNovel2018}
Patrick Helber, Benjamin Bischke, Andreas Dengel, and Damian Borth.
\newblock Introducing {Eurosat}: {A} {Novel} {Dataset} and {Deep} {Learning} {Benchmark} for {Land} {Use} and {Land} {Cover} {Classification}.
\newblock In \emph{{IGARSS} 2018 - 2018 {IEEE} {International} {Geoscience} and {Remote} {Sensing} {Symposium}}, pp.\  204--207, July 2018.
\newblock \doi{10.1109/IGARSS.2018.8519248}.
\newblock URL \url{https://ieeexplore.ieee.org/document/8519248}.

\bibitem[Hendrycks et~al.(2021{\natexlab{a}})Hendrycks, Basart, Mu, Kadavath, Wang, Dorundo, Desai, Zhu, Parajuli, Guo, Song, Steinhardt, and Gilmer]{hendrycksManyFacesRobustness2021}
Dan Hendrycks, Steven Basart, Norman Mu, Saurav Kadavath, Frank Wang, Evan Dorundo, Rahul Desai, Tyler Zhu, Samyak Parajuli, Mike Guo, Dawn Song, Jacob Steinhardt, and Justin Gilmer.
\newblock The {Many} {Faces} of {Robustness}: {A} {Critical} {Analysis} of {Out}-of-{Distribution} {Generalization}.
\newblock In \emph{Proceedings of the {IEEE}/{CVF} {International} {Conference} on {Computer} {Vision}}, pp.\  8340--8349, 2021{\natexlab{a}}.
\newblock URL \url{https://openaccess.thecvf.com/content/ICCV2021/html/Hendrycks_The_Many_Faces_of_Robustness_A_Critical_Analysis_of_Out-of-Distribution_ICCV_2021_paper.html}.

\bibitem[Hendrycks et~al.(2021{\natexlab{b}})Hendrycks, Zhao, Basart, Steinhardt, and Song]{hendrycksNaturalAdversarialExamples2021}
Dan Hendrycks, Kevin Zhao, Steven Basart, Jacob Steinhardt, and Dawn Song.
\newblock Natural {Adversarial} {Examples}.
\newblock In \emph{Proceedings of the {IEEE}/{CVF} {Conference} on {Computer} {Vision} and {Pattern} {Recognition}}, pp.\  15262--15271, 2021{\natexlab{b}}.
\newblock URL \url{https://openaccess.thecvf.com//content/CVPR2021/html/Hendrycks_Natural_Adversarial_Examples_CVPR_2021_paper.html}.

\bibitem[Hsieh et~al.(2023)Hsieh, Zhang, Ma, Kembhavi, and Krishna]{hsiehSugarCrepeFixingHackable2023}
Cheng-Yu Hsieh, Jieyu Zhang, Zixian Ma, Aniruddha Kembhavi, and Ranjay Krishna.
\newblock {SugarCrepe}: {Fixing} {Hackable} {Benchmarks} for {Vision}-{Language} {Compositionality}.
\newblock In \emph{Thirty-seventh {Conference} on {Neural} {Information} {Processing} {Systems} {Datasets} and {Benchmarks} {Track}}, November 2023.
\newblock URL \url{https://openreview.net/forum?id=Jsc7WSCZd4&noteId=Ekiryv85Mr}.

\bibitem[Islam et~al.(2024)Islam, Zaheer, Mahmood, and Nandakumar]{islamDiffuseMixLabelPreservingData2024}
Khawar Islam, Muhammad~Zaigham Zaheer, Arif Mahmood, and Karthik Nandakumar.
\newblock {DiffuseMix}: {Label}-{Preserving} {Data} {Augmentation} with {Diffusion} {Models}.
\newblock In \emph{Proceedings of the {IEEE}/{CVF} {Conference} on {Computer} {Vision} and {Pattern} {Recognition}}, pp.\  27621--27630, 2024.
\newblock URL \url{https://openaccess.thecvf.com/content/CVPR2024/html/Islam_DiffuseMix_Label-Preserving_Data_Augmentation_with_Diffusion_Models_CVPR_2024_paper.html}.

\bibitem[Kang et~al.(2023)Kang, Mun, Lee, and Roh]{Kang_2023_ICCV}
Wooyoung Kang, Jonghwan Mun, Sungjun Lee, and Byungseok Roh.
\newblock Noise-aware learning from web-crawled image-text data for image captioning.
\newblock In \emph{Proceedings of the IEEE/CVF International Conference on Computer Vision (ICCV)}, pp.\  2942--2952, October 2023.

\bibitem[Krause et~al.(2013)Krause, Stark, Deng, and Fei-Fei]{krause3DObjectRepresentations2013}
Jonathan Krause, Michael Stark, Jia Deng, and Li~Fei-Fei.
\newblock {3D} {Object} {Representations} for {Fine}-{Grained} {Categorization}.
\newblock In \emph{2013 {IEEE} {International} {Conference} on {Computer} {Vision} {Workshops}}, pp.\  554--561, December 2013.
\newblock \doi{10.1109/ICCVW.2013.77}.
\newblock URL \url{https://ieeexplore.ieee.org/document/6755945}.

\bibitem[Krizhevsky(2009)]{krizhevskyLearningMultipleLayers2009}
A.~Krizhevsky.
\newblock Learning {Multiple} {Layers} of {Features} from {Tiny} {Images}.
\newblock In \emph{Technical report}. University of Toronto, 2009.
\newblock URL \url{https://www.cs.toronto.edu/~kriz/learning-features-2009-TR.pdf}.

\bibitem[Kwon et~al.(2023)Kwon, Li, Zhuang, Sheng, Zheng, Yu, Gonzalez, Zhang, and Stoica]{kwonEfficientMemoryManagement2023}
Woosuk Kwon, Zhuohan Li, Siyuan Zhuang, Ying Sheng, Lianmin Zheng, Cody~Hao Yu, Joseph Gonzalez, Hao Zhang, and Ion Stoica.
\newblock Efficient {Memory} {Management} for {Large} {Language} {Model} {Serving} with {PagedAttention}.
\newblock In \emph{Proceedings of the 29th {Symposium} on {Operating} {Systems} {Principles}}, {SOSP} '23, pp.\  611--626, New York, NY, USA, October 2023. Association for Computing Machinery.
\newblock ISBN 9798400702297.
\newblock \doi{10.1145/3600006.3613165}.
\newblock URL \url{https://dl.acm.org/doi/10.1145/3600006.3613165}.

\bibitem[Lai et~al.(2024)Lai, Zhang, Zhang, Wu, Bai, Timofeev, Du, Gan, Shan, Chuah, Yang, and Cao]{laiVeCLIPImprovingCLIP2024}
Zhengfeng Lai, Haotian Zhang, Bowen Zhang, Wentao Wu, Haoping Bai, Aleksei Timofeev, Xianzhi Du, Zhe Gan, Jiulong Shan, Chen-Nee Chuah, Yinfei Yang, and Meng Cao.
\newblock {VeCLIP}: {Improving} {CLIP} {Training} via {Visual}-enriched {Captions}, March 2024.
\newblock URL \url{http://arxiv.org/abs/2310.07699}.

\bibitem[Li et~al.(2023{\natexlab{a}})Li, Wang, and Xie]{liCLIPAv2ScalingCLIP2023}
Xianhang Li, Zeyu Wang, and Cihang Xie.
\newblock {CLIPA}-v2: {Scaling} {CLIP} {Training} with 81.1\% {Zero}-shot {ImageNet} {Accuracy} within a \$10,000 {Budget}.
\newblock In \emph{R0-{FoMo}:{Robustness} of {Few}-shot and {Zero}-shot {Learning} in {Large} {Foundation} {Models}}, December 2023{\natexlab{a}}.
\newblock URL \url{https://openreview.net/forum?id=0hTtit3AAm}.

\bibitem[Li et~al.(2023{\natexlab{b}})Li, Wang, and Xie]{liInverseScalingLaw2023}
Xianhang Li, Zeyu Wang, and Cihang Xie.
\newblock An {Inverse} {Scaling} {Law} for {CLIP} {Training}.
\newblock In \emph{Thirty-seventh {Conference} on {Neural} {Information} {Processing} {Systems}}, November 2023{\natexlab{b}}.
\newblock URL \url{https://openreview.net/forum?id=LMU2RNwdh2}.

\bibitem[Li et~al.(2024)Li, Tu, Hui, Wang, Zhao, Xiao, Ren, Mei, Liu, Zheng, Zhou, and Xie]{liWhatIfWe2024}
Xianhang Li, Haoqin Tu, Mude Hui, Zeyu Wang, Bingchen Zhao, Junfei Xiao, Sucheng Ren, Jieru Mei, Qing Liu, Huangjie Zheng, Yuyin Zhou, and Cihang Xie.
\newblock What {If} {We} {Recaption} {Billions} of {Web} {Images} with {LLaMA}-3?, June 2024.
\newblock URL \url{http://arxiv.org/abs/2406.08478}.

\bibitem[Li et~al.(2023{\natexlab{c}})Li, Fan, Hu, Feichtenhofer, and He]{liScalingLanguageImagePreTraining2023}
Yanghao Li, Haoqi Fan, Ronghang Hu, Christoph Feichtenhofer, and Kaiming He.
\newblock Scaling {Language}-{Image} {Pre}-{Training} via {Masking}.
\newblock In \emph{Proceedings of the {IEEE}/{CVF} {Conference} on {Computer} {Vision} and {Pattern} {Recognition}}, pp.\  23390--23400, 2023{\natexlab{c}}.
\newblock URL \url{https://openaccess.thecvf.com/content/CVPR2023/html/Li_Scaling_Language-Image_Pre-Training_via_Masking_CVPR_2023_paper.html}.

\bibitem[Lian et~al.(2023)Lian, Li, Yala, and Darrell]{lianLLMgroundedDiffusionEnhancing2023}
Long Lian, Boyi Li, Adam Yala, and Trevor Darrell.
\newblock {LLM}-grounded {Diffusion}: {Enhancing} {Prompt} {Understanding} of {Text}-to-{Image} {Diffusion} {Models} with {Large} {Language} {Models}.
\newblock \emph{Transactions on Machine Learning Research}, October 2023.
\newblock ISSN 2835-8856.
\newblock URL \url{https://openreview.net/forum?id=hFALpTb4fR}.

\bibitem[Lin et~al.(2014)Lin, Maire, Belongie, Hays, Perona, Ramanan, Dollár, and Zitnick]{linMicrosoftCOCOCommon2014}
Tsung-Yi Lin, Michael Maire, Serge Belongie, James Hays, Pietro Perona, Deva Ramanan, Piotr Dollár, and C.~Lawrence Zitnick.
\newblock Microsoft {COCO}: {Common} {Objects} in {Context}.
\newblock In David Fleet, Tomas Pajdla, Bernt Schiele, and Tinne Tuytelaars (eds.), \emph{Computer {Vision} – {ECCV} 2014}, Lecture {Notes} in {Computer} {Science}, pp.\  740--755, Cham, 2014. Springer International Publishing.
\newblock ISBN 978-3-319-10602-1.
\newblock \doi{10.1007/978-3-319-10602-1_48}.

\bibitem[Liu et~al.(2019)Liu, Miao, Zhan, Wang, Gong, and Yu]{liuLargeScaleLongTailedRecognition2019a}
Ziwei Liu, Zhongqi Miao, Xiaohang Zhan, Jiayun Wang, Boqing Gong, and Stella~X. Yu.
\newblock Large-{Scale} {Long}-{Tailed} {Recognition} in an {Open} {World}.
\newblock In \emph{Proceedings of the {IEEE}/{CVF} {Conference} on {Computer} {Vision} and {Pattern} {Recognition} ({CVPR})}, pp.\  2537--2546, 2019.
\newblock URL \url{https://openaccess.thecvf.com/content_CVPR_2019/html/Liu_Large-Scale_Long-Tailed_Recognition_in_an_Open_World_CVPR_2019_paper.html}.

\bibitem[Loshchilov \& Hutter(2018)Loshchilov and Hutter]{loshchilovDecoupledWeightDecay2018}
Ilya Loshchilov and Frank Hutter.
\newblock Decoupled {Weight} {Decay} {Regularization}.
\newblock In \emph{International {Conference} on {Learning} {Representations}}, September 2018.
\newblock URL \url{https://openreview.net/forum?id=Bkg6RiCqY7}.

\bibitem[Loshchilov \& Hutter(2022)Loshchilov and Hutter]{loshchilovSGDRStochasticGradient2022}
Ilya Loshchilov and Frank Hutter.
\newblock {SGDR}: {Stochastic} {Gradient} {Descent} with {Warm} {Restarts}.
\newblock In \emph{International {Conference} on {Learning} {Representations}}, July 2022.
\newblock URL \url{https://openreview.net/forum?id=Skq89Scxx}.

\bibitem[Maji et~al.(2013)Maji, Rahtu, Kannala, Blaschko, and Vedaldi]{majiFineGrainedVisualClassification2013}
Subhransu Maji, Esa Rahtu, Juho Kannala, Matthew Blaschko, and Andrea Vedaldi.
\newblock Fine-{Grained} {Visual} {Classification} of {Aircraft}, June 2013.
\newblock URL \url{http://arxiv.org/abs/1306.5151}.

\bibitem[Mehta et~al.(2022)Mehta, Abdolhosseini, and Rastegari]{mehtaCVNetsHighPerformance2022}
Sachin Mehta, Farzad Abdolhosseini, and Mohammad Rastegari.
\newblock {CVNets}: {High} {Performance} {Library} for {Computer} {Vision}.
\newblock In \emph{Proceedings of the 30th {ACM} {International} {Conference} on {Multimedia}}, {MM} '22, pp.\  7327--7330, New York, NY, USA, October 2022. Association for Computing Machinery.
\newblock ISBN 978-1-4503-9203-7.
\newblock \doi{10.1145/3503161.3548540}.
\newblock URL \url{https://dl.acm.org/doi/10.1145/3503161.3548540}.

\bibitem[Mehta et~al.(2024{\natexlab{a}})Mehta, Abdolhosseini, and Rastegari]{AppleCorenet2024}
Sachin Mehta, Farzad Abdolhosseini, and Mohammad Rastegari.
\newblock apple/corenet, September 2024{\natexlab{a}}.
\newblock URL \url{https://github.com/apple/corenet}.

\bibitem[Mehta et~al.(2024{\natexlab{b}})Mehta, Horton, Faghri, Sekhavat, Najibi, Farajtabar, Tuzel, and Rastegari]{mehtaCatLIPCLIPlevelVisual2024}
Sachin Mehta, Maxwell Horton, Fartash Faghri, Mohammad~Hossein Sekhavat, Mahyar Najibi, Mehrdad Farajtabar, Oncel Tuzel, and Mohammad Rastegari.
\newblock {CatLIP}: {CLIP}-level {Visual} {Recognition} {Accuracy} with 2.7x {Faster} {Pre}-training on {Web}-scale {Image}-{Text} {Data}, April 2024{\natexlab{b}}.
\newblock URL \url{http://arxiv.org/abs/2404.15653}.

\bibitem[Mu et~al.(2022)Mu, Kirillov, Wagner, and Xie]{muSLIPSelfsupervisionMeets2022}
Norman Mu, Alexander Kirillov, David Wagner, and Saining Xie.
\newblock {SLIP}: {Self}-supervision {Meets} {Language}-{Image} {Pre}-training.
\newblock In \emph{Computer {Vision} – {ECCV} 2022: 17th {European} {Conference}, {Tel} {Aviv}, {Israel}, {October} 23–27, 2022, {Proceedings}, {Part} {XXVI}}, pp.\  529--544, Berlin, Heidelberg, October 2022. Springer-Verlag.
\newblock ISBN 978-3-031-19808-3.
\newblock \doi{10.1007/978-3-031-19809-0_30}.
\newblock URL \url{https://doi.org/10.1007/978-3-031-19809-0_30}.

\bibitem[Netzer et~al.(2011)Netzer, Wang, Coates, Bissacco, Wu, and Ng]{svhn}
Yuval Netzer, Tao Wang, Adam Coates, Alessandro Bissacco, Bo~Wu, and Andrew~Y. Ng.
\newblock Reading digits in natural images with unsupervised feature learning.
\newblock In \emph{NIPS Workshop on Deep Learning and Unsupervised Feature Learning 2011}, 2011.
\newblock URL \url{http://ufldl.stanford.edu/housenumbers/nips2011_housenumbers.pdf}.

\bibitem[Nilsback \& Zisserman(2008)Nilsback and Zisserman]{nilsbackAutomatedFlowerClassification2008}
Maria-Elena Nilsback and Andrew Zisserman.
\newblock Automated {Flower} {Classification} over a {Large} {Number} of {Classes}.
\newblock In \emph{2008 {Sixth} {Indian} {Conference} on {Computer} {Vision}, {Graphics} \& {Image} {Processing}}, pp.\  722--729, December 2008.
\newblock \doi{10.1109/ICVGIP.2008.47}.
\newblock URL \url{https://ieeexplore.ieee.org/document/4756141}.

\bibitem[OpenAI(2022)]{ChatGPT}
OpenAI.
\newblock Chatgpt, 2022.
\newblock URL \url{https://chatgpt.com}.

\bibitem[Parkhi et~al.(2012)Parkhi, Vedaldi, Zisserman, and Jawahar]{parkhiCatsDogs2012}
Omkar~M Parkhi, Andrea Vedaldi, Andrew Zisserman, and C.~V. Jawahar.
\newblock Cats and dogs.
\newblock In \emph{2012 {IEEE} {Conference} on {Computer} {Vision} and {Pattern} {Recognition}}, pp.\  3498--3505, June 2012.
\newblock \doi{10.1109/CVPR.2012.6248092}.
\newblock URL \url{https://ieeexplore.ieee.org/document/6248092}.

\bibitem[Plummer et~al.(2015)Plummer, Wang, Cervantes, Caicedo, Hockenmaier, and Lazebnik]{plummerFlickr30kEntitiesCollecting2015}
Bryan~A. Plummer, Liwei Wang, Chris~M. Cervantes, Juan~C. Caicedo, Julia Hockenmaier, and Svetlana Lazebnik.
\newblock Flickr30k {Entities}: {Collecting} {Region}-to-{Phrase} {Correspondences} for {Richer} {Image}-to-{Sentence} {Models}.
\newblock In \emph{2015 {IEEE} {International} {Conference} on {Computer} {Vision} ({ICCV})}, pp.\  2641--2649, December 2015.
\newblock \doi{10.1109/ICCV.2015.303}.

\bibitem[Podell et~al.(2024)Podell, English, Lacey, Blattmann, Dockhorn, M{\"u}ller, Penna, and Rombach]{podell2024sdxl}
Dustin Podell, Zion English, Kyle Lacey, Andreas Blattmann, Tim Dockhorn, Jonas M{\"u}ller, Joe Penna, and Robin Rombach.
\newblock {SDXL}: Improving latent diffusion models for high-resolution image synthesis.
\newblock In \emph{The Twelfth International Conference on Learning Representations}, 2024.
\newblock URL \url{https://openreview.net/forum?id=di52zR8xgf}.

\bibitem[Radford et~al.(2021)Radford, Kim, Hallacy, Ramesh, Goh, Agarwal, Sastry, Askell, Mishkin, Clark, Krueger, and Sutskever]{radfordLearningTransferableVisual2021}
Alec Radford, Jong~Wook Kim, Chris Hallacy, Aditya Ramesh, Gabriel Goh, Sandhini Agarwal, Girish Sastry, Amanda Askell, Pamela Mishkin, Jack Clark, Gretchen Krueger, and Ilya Sutskever.
\newblock Learning {Transferable} {Visual} {Models} {From} {Natural} {Language} {Supervision}.
\newblock In \emph{Proceedings of the 38th {International} {Conference} on {Machine} {Learning}}, pp.\  8748--8763. PMLR, July 2021.
\newblock URL \url{https://proceedings.mlr.press/v139/radford21a.html}.

\bibitem[Recht et~al.(2019)Recht, Roelofs, Schmidt, and Shankar]{rechtImageNetClassifiersGeneralize2019}
Benjamin Recht, Rebecca Roelofs, Ludwig Schmidt, and Vaishaal Shankar.
\newblock Do {ImageNet} {Classifiers} {Generalize} to {ImageNet}?
\newblock In \emph{Proceedings of the 36th {International} {Conference} on {Machine} {Learning}}, pp.\  5389--5400. PMLR, May 2019.
\newblock URL \url{https://proceedings.mlr.press/v97/recht19a.html}.

\bibitem[Rombach et~al.(2022)Rombach, Blattmann, Lorenz, Esser, and Ommer]{rombachHighResolutionImageSynthesis2022}
Robin Rombach, Andreas Blattmann, Dominik Lorenz, Patrick Esser, and Björn Ommer.
\newblock High-{Resolution} {Image} {Synthesis} {With} {Latent} {Diffusion} {Models}.
\newblock In \emph{Proceedings of the IEEE/CVF Conference on Computer Vision and Pattern Recognition (CVPR)}, pp.\  10684--10695, 2022.
\newblock URL \url{https://openaccess.thecvf.com/content/CVPR2022/html/Rombach_High-Resolution_Image_Synthesis_With_Latent_Diffusion_Models_CVPR_2022_paper.html}.

\bibitem[Schramowski et~al.(2023)Schramowski, Brack, Deiseroth, and Kersting]{schramowskiSafeLatentDiffusion2023}
Patrick Schramowski, Manuel Brack, Björn Deiseroth, and Kristian Kersting.
\newblock Safe {Latent} {Diffusion}: {Mitigating} {Inappropriate} {Degeneration} in {Diffusion} {Models}.
\newblock In \emph{Proceedings of the {IEEE}/{CVF} {Conference} on {Computer} {Vision} and {Pattern} {Recognition} ({CVPR})}, pp.\  22522--22531, 2023.
\newblock URL \url{https://openaccess.thecvf.com/content/CVPR2023/html/Schramowski_Safe_Latent_Diffusion_Mitigating_Inappropriate_Degeneration_in_Diffusion_Models_CVPR_2023_paper.html}.

\bibitem[Schuhmann et~al.(2022)Schuhmann, Beaumont, Vencu, Gordon, Wightman, Cherti, Coombes, Katta, Mullis, Wortsman, Schramowski, Kundurthy, Crowson, Schmidt, Kaczmarczyk, and Jitsev]{schuhmann2022laionb}
Christoph Schuhmann, Romain Beaumont, Richard Vencu, Cade~W Gordon, Ross Wightman, Mehdi Cherti, Theo Coombes, Aarush Katta, Clayton Mullis, Mitchell Wortsman, Patrick Schramowski, Srivatsa~R Kundurthy, Katherine Crowson, Ludwig Schmidt, Robert Kaczmarczyk, and Jenia Jitsev.
\newblock {LAION}-5b: An open large-scale dataset for training next generation image-text models.
\newblock In \emph{Thirty-sixth Conference on Neural Information Processing Systems Datasets and Benchmarks Track}, 2022.
\newblock URL \url{https://openreview.net/forum?id=M3Y74vmsMcY}.

\bibitem[Sharma et~al.(2018)Sharma, Ding, Goodman, and Soricut]{sharmaConceptualCaptionsCleaned2018}
Piyush Sharma, Nan Ding, Sebastian Goodman, and Radu Soricut.
\newblock Conceptual {Captions}: {A} {Cleaned}, {Hypernymed}, {Image} {Alt}-text {Dataset} {For} {Automatic} {Image} {Captioning}.
\newblock In Iryna Gurevych and Yusuke Miyao (eds.), \emph{Proceedings of the 56th {Annual} {Meeting} of the {Association} for {Computational} {Linguistics} ({Volume} 1: {Long} {Papers})}, pp.\  2556--2565, Melbourne, Australia, July 2018. Association for Computational Linguistics.
\newblock \doi{10.18653/v1/P18-1238}.
\newblock URL \url{https://aclanthology.org/P18-1238}.

\bibitem[Shi et~al.(2024)Shi, Wei, Zhou, Shao, Han, and Li]{shiLongTailLearningFoundation2024}
Jiang-Xin Shi, Tong Wei, Zhi Zhou, Jie-Jing Shao, Xin-Yan Han, and Yu-Feng Li.
\newblock Long-{Tail} {Learning} with {Foundation} {Model}: {Heavy} {Fine}-{Tuning} {Hurts}.
\newblock In \emph{Proceedings of the 41st {International} {Conference} on {Machine} {Learning}}, pp.\  45014--45039. PMLR, July 2024.
\newblock URL \url{https://proceedings.mlr.press/v235/shi24g.html}.

\bibitem[Stallkamp et~al.(2011)Stallkamp, Schlipsing, Salmen, and Igel]{stallkampGermanTrafficSign2011}
Johannes Stallkamp, Marc Schlipsing, Jan Salmen, and Christian Igel.
\newblock The {German} {Traffic} {Sign} {Recognition} {Benchmark}: {A} multi-class classification competition.
\newblock In \emph{The 2011 {International} {Joint} {Conference} on {Neural} {Networks}}, pp.\  1453--1460, July 2011.
\newblock \doi{10.1109/IJCNN.2011.6033395}.
\newblock URL \url{https://ieeexplore.ieee.org/document/6033395}.

\bibitem[Touvron et~al.(2023)Touvron, Martin, Stone, Albert, Almahairi, Babaei, Bashlykov, Batra, Bhargava, Bhosale, Bikel, Blecher, Ferrer, Chen, Cucurull, Esiobu, Fernandes, Fu, Fu, Fuller, Gao, Goswami, Goyal, Hartshorn, Hosseini, Hou, Inan, Kardas, Kerkez, Khabsa, Kloumann, Korenev, Koura, Lachaux, Lavril, Lee, Liskovich, Lu, Mao, Martinet, Mihaylov, Mishra, Molybog, Nie, Poulton, Reizenstein, Rungta, Saladi, Schelten, Silva, Smith, Subramanian, Tan, Tang, Taylor, Williams, Kuan, Xu, Yan, Zarov, Zhang, Fan, Kambadur, Narang, Rodriguez, Stojnic, Edunov, and Scialom]{touvronLlamaOpenFoundation2023}
Hugo Touvron, Louis Martin, Kevin Stone, Peter Albert, Amjad Almahairi, Yasmine Babaei, Nikolay Bashlykov, Soumya Batra, Prajjwal Bhargava, Shruti Bhosale, Dan Bikel, Lukas Blecher, Cristian~Canton Ferrer, Moya Chen, Guillem Cucurull, David Esiobu, Jude Fernandes, Jeremy Fu, Wenyin Fu, Brian Fuller, Cynthia Gao, Vedanuj Goswami, Naman Goyal, Anthony Hartshorn, Saghar Hosseini, Rui Hou, Hakan Inan, Marcin Kardas, Viktor Kerkez, Madian Khabsa, Isabel Kloumann, Artem Korenev, Punit~Singh Koura, Marie-Anne Lachaux, Thibaut Lavril, Jenya Lee, Diana Liskovich, Yinghai Lu, Yuning Mao, Xavier Martinet, Todor Mihaylov, Pushkar Mishra, Igor Molybog, Yixin Nie, Andrew Poulton, Jeremy Reizenstein, Rashi Rungta, Kalyan Saladi, Alan Schelten, Ruan Silva, Eric~Michael Smith, Ranjan Subramanian, Xiaoqing~Ellen Tan, Binh Tang, Ross Taylor, Adina Williams, Jian~Xiang Kuan, Puxin Xu, Zheng Yan, Iliyan Zarov, Yuchen Zhang, Angela Fan, Melanie Kambadur, Sharan Narang, Aurelien Rodriguez, Robert Stojnic, Sergey Edunov, and Thomas
  Scialom.
\newblock Llama 2: {Open} {Foundation} and {Fine}-{Tuned} {Chat} {Models}, July 2023.
\newblock URL \url{http://arxiv.org/abs/2307.09288}.

\bibitem[Trabucco et~al.(2023)Trabucco, Doherty, Gurinas, and Salakhutdinov]{trabuccoEffectiveDataAugmentation2023}
Brandon Trabucco, Kyle Doherty, Max~A. Gurinas, and Ruslan Salakhutdinov.
\newblock Effective {Data} {Augmentation} {With} {Diffusion} {Models}.
\newblock In \emph{The {Twelfth} {International} {Conference} on {Learning} {Representations}}, October 2023.
\newblock URL \url{https://openreview.net/forum?id=ZWzUA9zeAg}.

\bibitem[Udandarao et~al.(2024)Udandarao, Prabhu, Ghosh, Sharma, Torr, Bibi, Albanie, and Bethge]{udandaraoNoZeroShotExponential2024}
Vishaal Udandarao, Ameya Prabhu, Adhiraj Ghosh, Yash Sharma, Philip H.~S. Torr, Adel Bibi, Samuel Albanie, and Matthias Bethge.
\newblock No "{Zero}-{Shot}" {Without} {Exponential} {Data}: {Pretraining} {Concept} {Frequency} {Determines} {Multimodal} {Model} {Performance}, April 2024.
\newblock URL \url{http://arxiv.org/abs/2404.04125}.

\bibitem[Vasu et~al.(2024)Vasu, Pouransari, Faghri, and Tuzel]{vasuCLIPQualityCaptions2024}
Pavan Kumar~Anasosalu Vasu, Hadi Pouransari, Fartash Faghri, and Oncel Tuzel.
\newblock {CLIP} with {Quality} {Captions}: {A} {Strong} {Pretraining} for {Vision} {Tasks}, May 2024.
\newblock URL \url{http://arxiv.org/abs/2405.08911}.

\bibitem[Veeling et~al.(2018)Veeling, Linmans, Winkens, Cohen, and Welling]{veelingRotationEquivariantCNNs2018}
Bastiaan~S. Veeling, Jasper Linmans, Jim Winkens, Taco Cohen, and Max Welling.
\newblock Rotation {Equivariant} {CNNs} for {Digital} {Pathology}.
\newblock In \emph{Medical {Image} {Computing} and {Computer} {Assisted} {Intervention} – {MICCAI} 2018: 21st {International} {Conference}, {Granada}, {Spain}, {September} 16-20, 2018, {Proceedings}, {Part} {II}}, pp.\  210--218, Berlin, Heidelberg, September 2018. Springer-Verlag.
\newblock ISBN 978-3-030-00933-5.
\newblock \doi{10.1007/978-3-030-00934-2_24}.
\newblock URL \url{https://doi.org/10.1007/978-3-030-00934-2_24}.

\bibitem[von Platen et~al.(2022)von Platen, Patil, Lozhkov, Cuenca, Lambert, Rasul, Davaadorj, Nair, Paul, Berman, Xu, Liu, and Wolf]{von-platen-etal-2022-diffusers}
Patrick von Platen, Suraj Patil, Anton Lozhkov, Pedro Cuenca, Nathan Lambert, Kashif Rasul, Mishig Davaadorj, Dhruv Nair, Sayak Paul, William Berman, Yiyi Xu, Steven Liu, and Thomas Wolf.
\newblock Diffusers: {State}-of-the-art diffusion models, 2022.
\newblock URL \url{https://github.com/huggingface/diffusers}.

\bibitem[Wang et~al.(2019)Wang, Ge, Lipton, and Xing]{wangLearningRobustGlobal2019}
Haohan Wang, Songwei Ge, Zachary Lipton, and Eric~P Xing.
\newblock Learning {Robust} {Global} {Representations} by {Penalizing} {Local} {Predictive} {Power}.
\newblock In \emph{Advances in {Neural} {Information} {Processing} {Systems}}, volume~32. Curran Associates, Inc., 2019.
\newblock URL \url{https://proceedings.neurips.cc/paper/2019/hash/3eefceb8087e964f89c2d59e8a249915-Abstract.html}.

\bibitem[Wolf et~al.(2020)Wolf, Debut, Sanh, Chaumond, Delangue, Moi, Cistac, Rault, Louf, Funtowicz, Davison, Shleifer, von Platen, Ma, Jernite, Plu, Xu, Le~Scao, Gugger, Drame, Lhoest, and Rush]{wolfTransformersStateArtNatural2020}
Thomas Wolf, Lysandre Debut, Victor Sanh, Julien Chaumond, Clement Delangue, Anthony Moi, Pierric Cistac, Tim Rault, Remi Louf, Morgan Funtowicz, Joe Davison, Sam Shleifer, Patrick von Platen, Clara Ma, Yacine Jernite, Julien Plu, Canwen Xu, Teven Le~Scao, Sylvain Gugger, Mariama Drame, Quentin Lhoest, and Alexander Rush.
\newblock Transformers: {State}-of-the-{Art} {Natural} {Language} {Processing}.
\newblock In Qun Liu and David Schlangen (eds.), \emph{Proceedings of the 2020 {Conference} on {Empirical} {Methods} in {Natural} {Language} {Processing}: {System} {Demonstrations}}, pp.\  38--45, Online, October 2020. Association for Computational Linguistics.
\newblock \doi{10.18653/v1/2020.emnlp-demos.6}.
\newblock URL \url{https://aclanthology.org/2020.emnlp-demos.6}.

\bibitem[Xiao et~al.(2024)Xiao, Wu, Xu, Dai, Hu, Lu, Zeng, Liu, and Yuan]{xiaoFlorence2AdvancingUnified2024}
Bin Xiao, Haiping Wu, Weijian Xu, Xiyang Dai, Houdong Hu, Yumao Lu, Michael Zeng, Ce~Liu, and Lu~Yuan.
\newblock Florence-2: {Advancing} a {Unified} {Representation} for a {Variety} of {Vision} {Tasks}.
\newblock In \emph{Proceedings of the {IEEE}/{CVF} {Conference} on {Computer} {Vision} and {Pattern} {Recognition} ({CVPR})}, pp.\  4818--4829, 2024.
\newblock URL \url{https://openaccess.thecvf.com/content/CVPR2024/html/Xiao_Florence-2_Advancing_a_Unified_Representation_for_a_Variety_of_Vision_CVPR_2024_paper.html}.

\bibitem[Xiao et~al.(2010)Xiao, Hays, Ehinger, Oliva, and Torralba]{xiaoSUNDatabaseLargescale2010}
Jianxiong Xiao, James Hays, Krista~A. Ehinger, Aude Oliva, and Antonio Torralba.
\newblock {SUN} database: {Large}-scale scene recognition from abbey to zoo.
\newblock In \emph{2010 {IEEE} {Computer} {Society} {Conference} on {Computer} {Vision} and {Pattern} {Recognition}}, pp.\  3485--3492, June 2010.
\newblock \doi{10.1109/CVPR.2010.5539970}.
\newblock URL \url{https://ieeexplore.ieee.org/document/5539970}.

\bibitem[Yang et~al.(2024{\natexlab{a}})Yang, Yang, Hui, Zheng, Yu, Zhou, Li, Li, Liu, Huang, Dong, Wei, Lin, Tang, Wang, Yang, Tu, Zhang, Ma, Yang, Xu, Zhou, Bai, He, Lin, Dang, Lu, Chen, Yang, Li, Xue, Ni, Zhang, Wang, Peng, Men, Gao, Lin, Wang, Bai, Tan, Zhu, Li, Liu, Ge, Deng, Zhou, Ren, Zhang, Wei, Ren, Liu, Fan, Yao, Zhang, Wan, Chu, Liu, Cui, Zhang, Guo, and Fan]{yangQwen2TechnicalReport2024}
An~Yang, Baosong Yang, Binyuan Hui, Bo~Zheng, Bowen Yu, Chang Zhou, Chengpeng Li, Chengyuan Li, Dayiheng Liu, Fei Huang, Guanting Dong, Haoran Wei, Huan Lin, Jialong Tang, Jialin Wang, Jian Yang, Jianhong Tu, Jianwei Zhang, Jianxin Ma, Jianxin Yang, Jin Xu, Jingren Zhou, Jinze Bai, Jinzheng He, Junyang Lin, Kai Dang, Keming Lu, Keqin Chen, Kexin Yang, Mei Li, Mingfeng Xue, Na~Ni, Pei Zhang, Peng Wang, Ru~Peng, Rui Men, Ruize Gao, Runji Lin, Shijie Wang, Shuai Bai, Sinan Tan, Tianhang Zhu, Tianhao Li, Tianyu Liu, Wenbin Ge, Xiaodong Deng, Xiaohuan Zhou, Xingzhang Ren, Xinyu Zhang, Xipin Wei, Xuancheng Ren, Xuejing Liu, Yang Fan, Yang Yao, Yichang Zhang, Yu~Wan, Yunfei Chu, Yuqiong Liu, Zeyu Cui, Zhenru Zhang, Zhifang Guo, and Zhihao Fan.
\newblock Qwen2 {Technical} {Report}, July 2024{\natexlab{a}}.
\newblock URL \url{https://arxiv.org/abs/2407.10671v4}.

\bibitem[Yang et~al.(2024{\natexlab{b}})Yang, Yu, Meng, Xu, Ermon, and Cui]{yangMasteringTexttoImageDiffusion2024}
Ling Yang, Zhaochen Yu, Chenlin Meng, Minkai Xu, Stefano Ermon, and Bin Cui.
\newblock Mastering {Text}-to-{Image} {Diffusion}: {Recaptioning}, {Planning}, and {Generating} with {Multimodal} {LLMs}.
\newblock In \emph{Proceedings of the 41st {International} {Conference} on {Machine} {Learning}}, pp.\  56704--56721. PMLR, July 2024{\natexlab{b}}.
\newblock URL \url{https://proceedings.mlr.press/v235/yang24ai.html}.

\bibitem[Zhai et~al.(2020)Zhai, Puigcerver, Kolesnikov, Ruyssen, Riquelme, Lucic, Djolonga, Pinto, Neumann, Dosovitskiy, Beyer, Bachem, Tschannen, Michalski, Bousquet, Gelly, and Houlsby]{zhaiLargescaleStudyRepresentation2020}
Xiaohua Zhai, Joan Puigcerver, Alexander Kolesnikov, Pierre Ruyssen, Carlos Riquelme, Mario Lucic, Josip Djolonga, Andre~Susano Pinto, Maxim Neumann, Alexey Dosovitskiy, Lucas Beyer, Olivier Bachem, Michael Tschannen, Marcin Michalski, Olivier Bousquet, Sylvain Gelly, and Neil Houlsby.
\newblock A {Large}-scale {Study} of {Representation} {Learning} with the {Visual} {Task} {Adaptation} {Benchmark}, February 2020.
\newblock URL \url{http://arxiv.org/abs/1910.04867}.

\bibitem[Zhai et~al.(2022)Zhai, Wang, Mustafa, Steiner, Keysers, Kolesnikov, and Beyer]{zhaiLiTZeroShotTransfer2022}
Xiaohua Zhai, Xiao Wang, Basil Mustafa, Andreas Steiner, Daniel Keysers, Alexander Kolesnikov, and Lucas Beyer.
\newblock {LiT}: {Zero}-{Shot} {Transfer} {With} {Locked}-{Image} {Text} {Tuning}.
\newblock In \emph{Proceedings of the {IEEE}/{CVF} {Conference} on {Computer} {Vision} and {Pattern} {Recognition}}, pp.\  18123--18133, 2022.
\newblock URL \url{https://openaccess.thecvf.com/content/CVPR2022/html/Zhai_LiT_Zero-Shot_Transfer_With_Locked-Image_Text_Tuning_CVPR_2022_paper.html}.

\end{thebibliography}
\bibliographystyle{iclr2025_conference}

\appendix
\section{Appendix}
\subsection{Control Policies}
\label{sec:policy-details}
\paragraph{Text Prompt Templates.} We provide example control policies for text synthesis as predefined prompt templates, the first five templates do not include original text: 
\begin{enumerate}
\item "Create a detailed and high-quality caption using phrases that represent the entities or objects, their unique attributes, and the visual relationships in the scene depicted. Phrases: \{phrases\}."
\item "Compose a rich and immersive caption by incorporating a set of phrases that illustrate the entities or objects, their defining attributes, and the interconnections presented within the image. Phrases: \{phrases\}."
\item "Formulate an articulate and informative caption by using a series of phrases that outline the entities, their attributes, and their visual relationships depicted in an image. Phrases: \{phrases\}."
\item "Using a set of phrases that highlight the entities, attributes, and their visual associations in an image, craft a detailed and expressive caption. Phrases: \{phrases\}."
\item "Construct a comprehensive and expressive caption by integrating phrases that detail the entities, their features, and the spatial or thematic relationships in an image. Phrases: \{phrases\}."
\end{enumerate}
The following five templates include the original text, which is useful for maintaining the original meaning:
\begin{enumerate}
\item "Create a comprehensive caption that faithfully represents the objects, attributes, and their relationships contained within the provided sentence and phrases. Given sentence: \{caption\}. Given phrases: \{phrases\}. If the original caption specifies particular give phrases, maintain their integrity while using the phrases to enhance the description."
\item "Write a faithful caption by integrating the given phrases with the original sentence. Given sentence: \{caption\}. Given phrases: \{phrases\}. Ensure any objects or specific nouns from the original caption are preserved while elaborating on the visual relationships and attributes provided in the phrases to create a more detailed depiction."
\item "Provide a faithful and informative image caption using a given sentence and few phrases. Sentence: \{caption\}, phrases: \{phrases\}. Consider the initial sentence as a base for the overall context and ensure that specific objects or nouns such as numbers, car models, animals, etc., are preserved in the new caption. Integrate the given phrases, which describe entities, attributes, or visual relationships, to enrich and elaborate on the original meaning. Maintain fidelity to the original content while enhancing descriptive quality."
\item "Make a detailed caption based on the given phrases and a given sentence. Given phrases: \{phrases\}. Given sentence: \{caption\}. The sentence serves as a foundation, while the phrases elaborate on elements depicted in the image, like objects, their characteristics, and interactions. Preserve any pivotal information concerning objects, attributes, and their relations present in the sentence."
\item "Write a new faithful and high-quality caption based on the given phrases and a given sentence. The given sentence is the original caption and the phrases are entities or objects, attributes, and their visual relationships in an image. Given sentence: \{caption\}. Given phrases: \{phrases\}. If the sentence contains objects or nouns (e.g. digits, car models, planes, pets, animals, etc.), the new caption should be faithful and keep this information. Otherwise, use the phrases to create the new caption."
\end{enumerate}

\paragraph{Image Prompt Templates.} We provide five image prompt templates:
\begin{enumerate}
\item "real": "a real photo. \{prompt\}. 35mm photograph, film, bokeh, professional, 4k, highly detailed",
\item "nocap": "a real photo showing \{prompt\}. highly detailed"
\item "isometric": "isometric style \{prompt\} . vibrant, beautiful, crisp, detailed, ultra detailed, intricate"
\item "enhance": "breathtaking \{prompt\}. award-winning, professional, highly detailed"
\item "quality": "masterpiece, best quality, ultra detailed, \{prompt\}. intricate details"
\end{enumerate}

\subsection{Datasets Details}
\label{sec:dataset-details}
\paragraph{Evaluation Datasets.} We list the vision datasets for evaluation in \cref{tab:datasets}.

\begin{table}[ht!]
 \centering
\caption{\small Details of evaluation datasets.}\label{tab:datasets}
\small
\begin{tabular}{@{}llrr@{}}
\toprule
Dataset                & Metric                &  Classes & Test Set Size \\ \midrule
CIFAR-10~\citep{krizhevskyLearningMultipleLayers2009}               & Accuracy              & 10                & 10000         \\
CIFAR-100~\citep{krizhevskyLearningMultipleLayers2009}              & Accuracy              & 100               & 10000         \\
CLEVR Counts           & Accuracy              & 8                 & 15000         \\
CLEVR Distance         & Accuracy              & 6                 & 15000         \\
Caltech-101~\citep{fei-feiOneshotLearningObject2006}            & Mean Per Class Recall & 102               & 6085          \\
Country211~\citep{radfordLearningTransferableVisual2021}             & Accuracy              & 211               & 21100         \\
DTD~\citep{cimpoiDescribingTexturesWild2014}   & Accuracy              & 47                & 1880          \\
EuroSAT~\citep{helberIntroducingEurosatNovel2018}                & Accuracy              & 10                & 5400          \\
FGVC Aircraft~\citep{majiFineGrainedVisualClassification2013}          & Mean Per Class Recall & 100               & 3333          \\
Food-101~\citep{bossardFood101MiningDiscriminative2014}               & Accuracy              & 101               & 25250         \\
GTSRB ~\citep{stallkampGermanTrafficSign2011}                 & Accuracy              & 43                & 12630         \\
KITTI~\citep{geigerVisionMeetsRobotics2013}  & Accuracy              & 4                 & 711           \\
Oxford Flowers-102~\citep{nilsbackAutomatedFlowerClassification2008}     & Mean Per Class Recall & 102               & 6149          \\
Oxford-IIIT Pet~\citep{parkhiCatsDogs2012}        & Mean Per Class Recall & 37                & 3669          \\
PatchCamelyon~\citep{veelingRotationEquivariantCNNs2018}          & Accuracy              & 2                 & 32768         \\
RESISC45~\citep{chengRemoteSensingImage2017}               & Accuracy              & 45                & 6300          \\
STL-10~\citep{coatesAnalysisSingleLayerNetworks2011}                 & Accuracy              & 10                & 8000          \\
SUN397~\citep{xiaoSUNDatabaseLargescale2010}                & Accuracy              & 397               & 108754        \\
SVHN~\citep{svhn}                  & Accuracy              & 10                & 26032         \\
Stanford Cars~\citep{krause3DObjectRepresentations2013}          & Accuracy              & 196               & 8041          \\
ImageNet-1K~\citep{dengImageNetLargescaleHierarchical2009}           & Accuracy              & 1000              & 50000         \\
ImageNet-V2~\citep{rechtImageNetClassifiersGeneralize2019}           & Accuracy              & 1000              & 10000         \\
ImageNet-S~\citep{wangLearningRobustGlobal2019}            & Accuracy              & 1000              & 50889         \\
ImageNet-A~\citep{hendrycksNaturalAdversarialExamples2021}            & Accuracy              & 200               & 7500          \\
ImageNet-O~\citep{hendrycksNaturalAdversarialExamples2021}            & Accuracy              & 200               & 2000          \\
ImageNet-R~\citep{hendrycksManyFacesRobustness2021}            & Accuracy              & 200               & 30000         \\

Flickr~\citep{plummerFlickr30kEntitiesCollecting2015}                 & Mean Recall@1         & -                 & 1000          \\
MSCOCO ~\citep{linMicrosoftCOCOCommon2014}                & Mean Recall@1         & -                 & 5000          \\ \bottomrule
\end{tabular}
\end{table}

\paragraph{Long-tail Datasets.} For the tail classes in ImageNet-LT and Places-LT, we generate synthetic images using the ``real'' style of image prompt template, and we generate 7 samples per tail class so that we roughly double the size of the original real datasets. We obtain 80.4k synthetic samples for ImageNet-LT and 55.2K for Places-LT.

\subsection{Training Details}
\label{sec:training-details}

\begin{table*}[t!]
    \centering
    \caption{Training hyper-parameters.}
    \label{tab:training-hparam}
    \begin{subtable}[t]{0.49\columnwidth}
        \centering
        \caption{Pretraining CLIP on CC3M and CC12M.}
        \label{tab:pretrain-cc3m}
        \resizebox{!}{60px}{
        \begin{tabular}{lcc}
            \toprule[1.5pt]
            \textbf{Hyperparameter} & \textbf{CC3M} & \textbf{CC12M} \\
            \midrule[1pt]
            Total iterations & 56,429 & 55,429 \\
            Warmup iterations & 2822  & 2771 \\
            Image size & 224 & 224 \\
            LR scheduler & Cosine & Cosine \\
            Max. LR & 0.002 & 0.002 \\
            Min. LR & 0.00002 & 0.00002 \\
            Optimizer & AdamW & AdamW \\
            AdamW $\beta$'s & (0.9, 0.98) & (0.9, 0.98) \\
            Weight decay & 0.2 & 0.2 \\
            Batch size per GPU & 256 & 256 \\
            \# A100 GPUs & 8 & 32 \\
            A100 GPU Memory & 40 GB & 40 GB \\
            \bottomrule[1.5pt]
        \end{tabular}
        }
    \end{subtable}
    \hfill
    \begin{subtable}[t]{0.49\columnwidth}
        \centering
        \caption{Finetuning CLIP on Places-LT and ImageNet-LT.}
        \label{tab:ft-longtail}
        \resizebox{!}{60px}{
        \begin{tabular}{lcc}
            \toprule[1.5pt]
            \textbf{Hyperparameter} & \textbf{Places-LT} & \textbf{ImageNet-LT} \\
            \midrule[1pt]
            Total Iterations & 56,429 & 55,429 \\
            Warmup Iterations & 2822  & 2771 \\
            Image size & 224 & 224 \\
            Loss type & CrossEntropy & CrossEntropy \\
            LR scheduler & Cosine & Cosine \\
            Learning rate &  0.01 &  0.01 \\
            Optimizer & SGD & SGD \\
            Momentum & 0.9 & 0.9 \\
            Weight decay & 5e-4 & 5e-4 \\
            Batch size per GPU & 128 & 128 \\
            \# A100 GPUs & 1 & 1 \\
            A100 GPU Memory & 40 GB & 40 GB \\
            \bottomrule[1.5pt]
        \end{tabular}
        }
    \end{subtable}
    \vskip -0.1in
\end{table*}

\paragraph{Pretraining Hyper-parameters.} We pretrain the CLIP for the same number of iterations for both the baseline and \sys. For example, suppose we train for $E$ epochs, if the original dataset has $N$ samples, \sys has generated $N^\prime$ samples ($N^\prime <= N$ due to filtering), then the total samples are $E*N$, we train \sys models for $\frac{E*N}{N+N^\prime}$ epochs. This guarantees that the baseline and \sys CLIP models have seen the same number of data samples. 

\cref{tab:training-hparam} lists the hyper-parameters used for pretraining on CC3M and CC12m.  We use AdamW~\citep{loshchilovDecoupledWeightDecay2018} with default $\beta$ values as an optimizer and binary cross-entropy loss as an objective function. We use cosine learning rate schedule~\citep{loshchilovSGDRStochasticGradient2022}. We use the CoreNet library~\citep{AppleCorenet2024,mehtaCVNetsHighPerformance2022} for all pretraining experiments. We adapt the LIFT codebase~\citep{shiLongTailLearningFoundation2024} for fine-tuning long-tail tasks, main modifications include adding support for iteration-based training and data loader for multiple datasets.

\subsection{\sys Inference Details}
\label{sec:inference-details}
\paragraph{VTM.} We use a hybrid tagging model consisting of two stages. We first run the ViT-Huge variant of CatLIP~\citep{mehtaCatLIPCLIPlevelVisual2024} for each image and output top20 classes based on the sigmoid score of prediction logits, then we convert the class indices to actual word labels. The vocabulary size of CatLIP is 24320. Most of the vocabulary words are nouns and single-word attributes. 
We then run the Florence-large~\citep{xiaoFlorence2AdvancingUnified2024} for each image to extract detailed captions using the task prompt {\texttt{<MORE_DETAILED_CAPTION>}}. After that, we run Qwen2-7B-Instruct~\citep{yangQwen2TechnicalReport2024} to extract objects, attributes, and relations from the Florence captions. We then merge the objects field with CatLIP-predicted labels. The extraction instruction contains a 2-shot example and we list the prompt template below:
\begin{lstlisting}
For a given image caption, identify all the attributes, objects or entities, and visual relationships or actions that are phrases. The phrases should only come from the caption. Separate the phrases by comma without formatting. Output three lines:
attributes: phrases
objects: phrases
relations: phrases

Examples:

caption: The image is a close-up portrait of a middle-aged man wearing a white cowboy hat. He appears to be in his late 60s or early 70s, with gray hair and a serious expression on his face. He is wearing a dark suit jacket and a light blue collared shirt. The background is a clear blue sky with trees visible in the distance. The man is looking off to the side with a slight smile on his lips.
attributes: close-up, middle-aged, white cowboy hat, gray hair, serious expression, light blue
objects: portrait, man, hat, face, dark suit jacket, shirt, blue sky, trees, lips
relations: wearing a, visible in the distance, looking off to the side, slight smile on his lips

caption: The image shows a female singer performing on a stage. She is standing on a set of stairs with her legs spread apart and holding a microphone in her hand. The stage is lit up with red and blue lights and there is a large circular screen in the background. The singer is wearing a black and white patterned outfit with high heels. She appears to be in the middle of a song or performance.
attributes: female singer, stage, set of stairs, red and blue lights, large circular screen, black and white patterned outfit, high heels
objects: female singer, stage, set of stairs, legs, microphone, screen, outfit, high heels, song, performance
relations: performing on a stage, standing on, her legs spread apart, holding, lit up, background, wearing, in the middle of a song

caption: {caption}
\end{lstlisting}

CatLIP is available in CoreNet so we use it directly for inference and we wrap the Florence Transformers~\citep{wolfTransformersStateArtNatural2020} code into the CoreNet inference pipeline for easier integration.

\paragraph{LLM.} We use the vLLM engine~\citep{kwonEfficientMemoryManagement2023} for offline inference in Qwen2 and Mistral-Nemo. We use greedy decoding for the generation.

\paragraph{Text-to-image Model.} We use the diffusers~\citep{von-platen-etal-2022-diffusers} library for diffusion model inference. For both SDXL and SD3M models, we use float16 dtype with a guidance scale of 7.0 and set the diffusion steps to 28.

\subsection{More Analysis Details}
\label{sec:analysis-details}

\paragraph{\sys Samples.} For CC3M, the original dataset has 2.8 million image-caption pairs, \sys-cap contains 2.6 million captions, \sys-img contains 2.4 million images, and \sys-mix contains 5.1 million image-caption pairs. Original CC12M has 11.3 million image-caption samples, \sys-cap consists of 10.2 million captions, \sys-img contains 9.5 million images, and \sys-mix has 19.7 million image-caption pairs.

\begin{figure}
    \centering
    \includegraphics[width=0.5\linewidth]{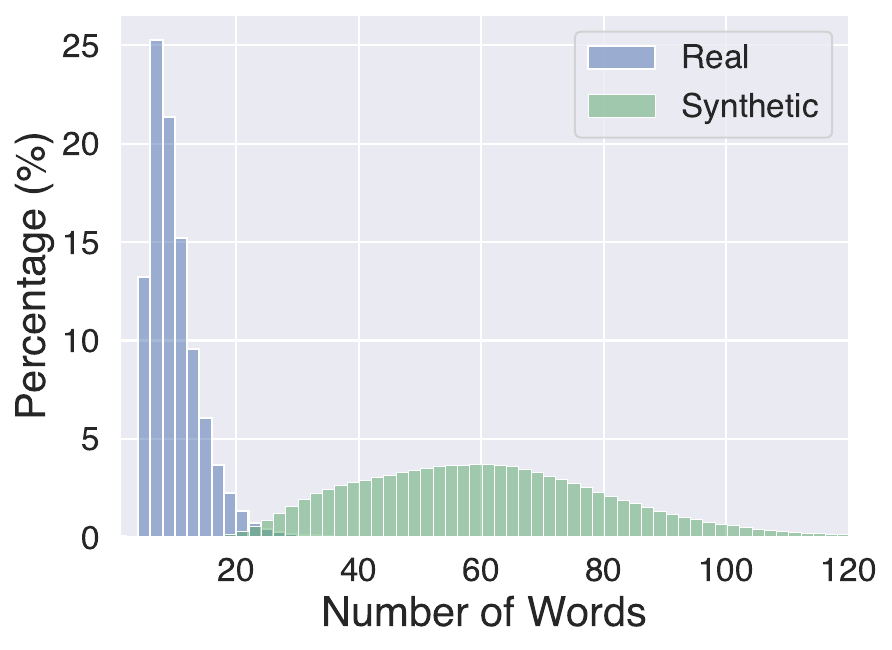}
    \caption{\small{Number of words for the original captions and \sys synthetic texts on CC3M.}}
    \label{fig:words-freq}
\end{figure}

\paragraph{\sys Synthetic Texts.} We plot the number of words for synthetic texts generated by \sys and compare them with original real texts in \cref{fig:words-freq}.

\end{document}